\documentclass[sensors,preprints,submit,moreauthors,pdftex,10pt,a4paper]{mdpi}

\pdfoutput=1

\preto{\abstractkeywords}{\nolinenumbers}

\firstpage{1} 

\usepackage{booktabs}
\usepackage{multirow}

\usepackage{moreverb,url}
\usepackage{acronym}
\usepackage{gensymb}
\usepackage{mathtools}
\usepackage{amsmath}
\usepackage{mleftright}	
\usepackage{bigints}
\usepackage{relsize}
\usepackage[scientific-notation=true]{siunitx}
\usepackage[labelformat=simple]{subcaption}

\DeclareCaptionLabelFormat{subcaptionlabel}{\normalfont(\textbf{#2}\normalfont)}
\usepackage{threeparttable}
\usepackage{tabulary}
\usepackage{xcolor}
\usepackage{graphicx}
\usepackage{bbm}

\newacro{DOF}{degree of freedom}
\newacro{RTK} {real time kinematic}
\newacro{GPS} {global positioning system}
\newacro{INS} {inertial navigation system}
\newacro{ACFR} {Australian Centre for Field Robotics}
\newacro{MC} {Monte Carlo}
\newacro{DEM} {digital elevation model}
\newacro{GCP} {ground control point}
\newacro{MCMC} {Markov Chain Monte Carlo}
\newacro{UAV}{unmanned aerial vehicle}
\newacro{IFOV}{instantaneous field of view}
\newacro{IMU}{inertial measurement unit}
\newacro{KDE}{kernel density estimate}
\newacro{VNIR}{visible to near infrared}

\newcommand{\argmin}[1]{\underset{#1}{argmin}\;}

\Title{Extrinsic Parameter Calibration for Line Scanning Cameras on Ground Vehicles with Navigation Systems Using a Calibration Pattern}

\Author{Alexander Wendel * and James Underwood}
\AuthorNames{Alexander Wendel and James Underwood}

\address[1]{
The Australian Centre for Field Robotics (ACFR), Department of Aerospace, Mechanical and Mechatronic Engineering (AMME), The University of Sydney, Sydney, NSW 2006, Australia; j.underwood@acfr.usyd.edu.au} 

\corres{\hangafter=1 \hangindent=1.05em \hspace{-0.82em} Correspondence: a.wendel@acfr.usyd.edu.au; Tel.: +61 2 9114 0895 } 

\abstract{Line scanning cameras, which capture only a single line of pixels, have been increasingly used in ground based mobile or robotic platforms. In applications where it is advantageous to directly georeference the camera data to world coordinates, an accurate estimate of the camera's 6D pose is required. This paper focuses on the common case where a mobile platform is equipped with a rigidly mounted line scanning camera, whose pose is unknown, and a navigation system providing vehicle body pose estimates. We propose a novel method that estimates the camera's pose relative to the navigation system. The approach involves imaging and manually labelling a calibration pattern with distinctly identifiable points, triangulating these points from camera and navigation system data and reprojecting them in order to compute a likelihood, which is maximised to estimate the 6D camera pose. Additionally, a Markov Chain Monte Carlo (MCMC) algorithm is used to estimate the uncertainty of the offset. Tested on two different platforms, the method was able to estimate the pose to within 0.06 m/1.05\degree  \,and 0.18 m/2.39\degree. We also propose several approaches to displaying and interpreting the 6D results in a human readable way.}


\begin{document}

\section{Introduction}
Line scanning (also 1D or linear) cameras, which produce a single line of pixels for each exposure, have been used widely in areas such as remote sensing \citep{yang2004airborne,bethel2000geometric} and industrial inspection \citep{lim2008visual,dale2013,pfaff2017real}. While 2D frame cameras offer the benefit of imaging a larger scene with each exposure, linescan cameras allow capturing of images at higher frame rates or spatial resolution \citep{li2016cross}. One specific but common example is hyperspectral line scanning cameras, which provide both high spatial and spectral resolution. Many~applications require accurate and direct determination of the real world coordinates of line scan image data, also known as georeferencing or mapping. This requires precise calibration of the sensor's intrinsic (e.g., focal length and principal point) and extrinsic parameters (i.e., camera pose with respect to the vehicle body frame). In the remote sensing literature, determination of extrinsic parameters is known as lever arm (translation) and boresight (orientation) alignment. More recently line scanning cameras have also been studied for low altitude \ac{UAV} and mobile ground based applications \citep{ramirez2015low,deery2014proximal,wendelunderwoodweeddetection,trierscheid2008hyperspectral}, but there are fewer studies addressing the extrinsic calibration requirements that closer proximity to the scene implies. Requirements include obtaining a 6 \ac{DOF} extrinsic parameter solution including translation, which has a greater influence on mapping when proximal; avoiding \acp{GCP}, which need to be more accurately geolocated when viewed from nearby; and a need for smaller survey areas for calibration, because it is more difficult to obtain data over large areas with mobile ground vehicles. This paper addresses these requirements by providing a novel method to estimate line scanning camera pose with respect to the platform body frame, where the location and orientation of the platform is itself provided in world coordinates from a navigation system. The method uses the data from the navigation and line scanning camera only, avoiding the need for auxiliary sensors.

Extrinsic calibration for 2D frame cameras has been studied extensively due to their ubiquitous use across many different fields, and established solutions exist \citep{zhang2000flexible, mostafa2001boresight,lobo2007relative,hol2010modeling}.  Calibration of 1D cameras has not received as much attention. Methods can be loosely grouped into two categories: scan-based calibration and line-based calibration \citep{li2016cross}. Scan-based calibration requires an accurate rig with a linear actuator that moves the camera orthogonally to the line scan at a constant speed over a calibration pattern, such as a checker board \citep{drareni2011plane,hui2012line}. This method is suitable for industrial inspection applications in a controlled laboratory or factory setting, where a linear actuator, manipulator arm or other rig is capable of moving the sensor through a precisely specified trajectory. Line-based calibration methods, on the other hand, allow calibration from a single line scan of a 3D target with a carefully designed pattern of lines \citep{li2014cross,luna2010calibration}. Line-based approaches require that the dimensions of the calibration pattern are known precisely, and that the whole pattern has been imaged in one exposure. Recently, a variation of this method using multiple line scans of a planar calibration pattern has been proposed \citep{yao2014geometric}, and~the use of an additional auxiliary frame camera has also been explored \citep{li2016cross,sun2016calibration}. All the aforementioned approaches are suitable for well controlled environments: for scan-based calibration the movement of the sensor needs to be accurately controlled, while for line-based methods, the position of the pattern \textls[-25]{with respect to the sensor is critical. However, in a mobile ground based field platform, where the camera is rigidly mounted in a particular position to the platform, it is difficult to meet either of those requirements.}

In previous methods, extrinsic parameters are usually determined with respect to the calibration pattern or an auxiliary frame camera. Therefore to determine the camera to navigation system transform either requires accurate knowledge of a pattern or points in world coordinates or an additional step such as ``hand-eye'' calibration \citep{Ma2014}. Hand-eye calibration involves determining the transformation from a camera to an end effector (a robotic hand for instance), where these are rigidly linked, and is a thoroughly covered topic in the robotics literature. The problem is generally solved by imaging a calibration pattern from many different locations, where the transformations between the different end effector positions and camera to calibration pattern transformations are known using standard frame camera calibration techniques. Comparisons can be made with the problem in this paper, where the navigation system positions (and therefore any transformations between them) are known, and camera to calibration pattern transformations can be determined using any of the previously discussed methods. 

\textls[-15]{As remote sensing most commonly involves imaging from an aerial or satellite platform, translation (lever arm) offsets have a smaller effect on imaging accuracy, and can be measured manually} \citep{6947103,perry2008precision}. Accurately geolocated \acp{GCP} are commonly used to determine boresight alignment \citep{muller2002program}, which can also be adopted for ground based applications \citep{abd2016georeferencing}. Efforts have been made to avoid the use of \acp{GCP}, by detecting points of interest in separate scans of the same area and determining their 3D position using a known \ac{DEM} \citep{6947103}. Similarly, non-surveyed tie-points between overlapping acquisition runs have been used in combination with bundle adjustment to determine boresight parameters \citep{wohlfeil2009modular}. The use of \acp{GCP} has also been combined with \acp{DEM} to improve accuracy and allow self-calibration \citep{yeh2011self}. Frame cameras have been used to aid in determination of boresight misalignments~\citep{wohlfeil2012optical}, and additionally in combination with a \ac{DEM} \citep{barbieux2016correction}. Frame camera images have also been used to improve the geometric characteristics of processed hyperspectral linescan images from a \ac{UAV} \citep{7409968}.

This paper provides a method for the determination of the relative 6 \ac{DOF} pose of a rigidly mounted line scanning camera with respect to a navigation system on a ground based mobile platform. With this approach many of the previously outlined requirements and limitations are mitigated:

\begin{itemize}[leftmargin=*,labelsep=6mm]
	\item The dimensions of the calibration pattern do not need to be known, and so it does not need to be printed to any particular accuracy, nor even measured.
	\item \acp{GCP} do not need to be surveyed.
	\item Auxiliary sensors, such as 2D frame cameras, are not required to aid the calibration.
	\item A single, compact calibration pattern can be used rather than widely distributed \acp{GCP}.
	\item Translational (lever arm) offsets are determined in addition to rotations (boresight), due to their increased significance when at close proximity to the scene.
\end{itemize}

The remainder of the paper is organised as follows. In Section \ref{sec:general_approach} the theory of the proposed method is outlined in detail. Then Section \ref{sec:methods} provides practical implementation details and the experimental method. Experimental results using the Ladybird and Shrimp robotic platforms are produced and discussed in Sections \ref{sec:results} and \ref{sec:discussion}.

\section{Overview of Approach} \label{sec:general_approach}

In this section, the theoretical approach used for estimating the camera pose with respect to the platform body is outlined in detail. Initially, an overview of the line scanning camera model is provided, which is an adaptation of the widely used pinhole model. This allows defining lines or rays in 3D space that intersect both the camera centre and a pixel on the sensor. When combined with the Cartesian coordinate transformations between camera, body and world frames, rays can be projected onto a surface, and conversely a world 3D point can be reprojected to a point on the 2D sensor. It is desirable to minimise any errors in the camera pose, as they directly affect mapping accuracy. 

We propose a method that estimates the relative camera pose using image and navigation system data. The data are obtained by moving the platform in order to observe a calibration pattern with multiple point targets from different perspectives. The calibration pattern point locations are then manually labelled in the image data. Starting from an initial hand measured camera pose, image pixel locations of the observed pattern points and corresponding platform poses are combined, and all of the resulting rays are used to triangulate the pattern point locations in world coordinates. These point estimates are then reprojected to the sensor frame for each observation. The reprojection error is calculated as the distance between each observed and reprojected pixel. The reprojection error uncertainty is calculated by propagating the input uncertainties through each calculation as variance-covariance matrices (henceforth referred to as covariance matrices for brevity). Assuming a normal distribution of the reprojection error over input parameters, the likelihood of the data given a relative camera pose hypothesis can be estimated. By maximising the likelihood, the six relative camera pose parameters can be optimised. Following this, a random sampling based procedure is provided to estimate the uncertainties of the optimal camera pose using \ac{MCMC}.

Throughout this paper, superscripts represent the reference frame of a particular variable. Subscripts refer to a descriptor (e.g., which pose is being referred to), axis reference, and instance identifiers for that variable, in that order. For example, $r_{c,x}^b$ refers to the camera centre location along the $x$ axis relative to the body frame.

\subsection{Line Scanning Camera Model} \label{sec:1d_cam_model}
Using the pinhole camera model with homogeneous coordinates, a point $\mathbf {p}^w = [x, y, z, 1]^T$ in world coordinates is projected to the camera sensor at $[u, v, 1]^T$ with the following Equation \citep{hartley2003multiple}:

\begin{equation} \label{sec:pinhole_projection}
\left( \begin{array}{c} u \\ v \\ 1 \end{array} \right)s = 
\mathbf{P} \left( \begin{array}{c} x \\ y \\ z \\ 1 \end{array} \right),
\end{equation}
where $s$ is a scale factor and $\textbf{P}$ can be broken down into,

\begin{equation} \label{eq:P}
\mathbf{P} = \mathbf{K} {\mathbf{R}_c^w}^{-1} [\mathbf{I}_{3 \times 3} |-\mathbf{p}^w_c ].
\end{equation}

$\mathbf{R}_c^w$ is the rotation matrix of the camera with respect to the world frame. Joined horizontally are $\mathbf{I}_{3 \times 3}$ and $\mathbf{p}^w_c$, which are the identity matrix and the world camera position (i.e., the camera centre $[r_{c,x}^w, r_{c,y}^w, r_{c,z}^w]^T$) respectively.  $\mathbf{K}$ is the intrinsic camera matrix:

\begin{equation}
\mathbf{K} = \left[ \begin{array}{ccc} f & 0 & u_0 \\ 0 & f & v_0 \\ 0 & 0 & 1 \end{array} \right],
\end{equation}
where $f$, $u_0$ and $v_0$ are the focal length (in pixels) and principal points respectively (we neglect skew because there is only one spatial axis). For a line scanning camera, we assume that $v_0 = 0$ and so it follows that for a 3D world point to be visible in the 1D pixel array, it must be located near the plane that intersects the scan line on the sensor (i.e., where $v = 0$) and the camera centre (focal point). How~closely a point must be located to that plane depends on the \ac{IFOV} and distance from the sensor. The \ac{IFOV} is the angle over which each pixel is sensitive to radiation. While linescan image data is by definition at $v = 0$, reprojection errors can occur in both $u$ and $v$ as will be shown later. Therefore, even though the model allows for two spatial dimensions on the image sensor, it describes the projection of points for individual 1D line scan frames only. 

Each pixel point $[u, v, 1]^T$ maps to a ray or line in 3D space, which connects the sensor pixel, camera centre and object being viewed. While that ray may be defined by any two points that lie on it, the following are mathematically convenient to obtain: the camera centre $\mathbf{p}^w_c$ and $\mathbf{p}^w_s = \mathbf{P}^+ [u, v, 1]^T$, where $\mathbf{P}^+$ is the pseudo-inverse of $\mathbf{P}$ \citep{hartley2003multiple}. 

\subsection{Rotation and Transform Conventions} \label{sec:rotation_matrix}
In this paper, we use both Euler and axis-angle conventions to represent rotations compactly. The~navigation system on the platforms used in this work provide platform pose estimates using the Euler $zyx$ intrinsic convention (also known as Tait-Bryan or yaw-pitch-roll), which are represented~as~$[\phi_{x},\phi_{y},\phi_{z}]$, and may be converted to rotation matrices as per \citet{berner2007orientation} or Section \ref{sec3.1} in \citet{underwood2009reliable}. While~Euler angle representations are commonly used in robotics applications, they present the following ambiguities. Some different combinations of $\phi_{x}$ $\phi_{y}$ and $\phi_{z}$ can represent the same rotation~\citep{Huynh2009}. Similarly, a small freedom of rotation about a non-orthogonal axis can result in a large correlated degree of freedom spread over two Euler angles, which is difficult to interpret when estimating parameter uncertainty. For these reasons, while navigation and hand measured pose data is provided as Euler angles, we favour the axis-angle representation for all internal calculations and results. An axis-angle rotation is given as a unit length vector $\mathbf{e}$ and a rotation $\theta$ around it:

\begin{equation}
\left(\theta, \mathbf{e}\right) = \left( \theta, \begin{bmatrix} e_x \\ e_y \\ e_z \end{bmatrix} \right).
\end{equation}

Since rotations only have three degrees of freedom, an axis-angle rotation may be expressed as a length three vector:

\begin{equation}
\theta \mathbf{e} = \begin{bmatrix} \theta e_x \\ \theta e_y \\ \theta e_z \end{bmatrix}.
\end{equation}

Axis-angle rotations may be converted to rotation matrices as follows \citep{berner2007orientation}:

\begin{equation}
\mathbf{R} = \left[ \mathbf{I}_{3x3} + sin(\theta)\mathbf{S}_n + (1-cos(\theta)) \mathbf{S}_n^2 \right],
\end{equation}
where

\begin{equation}
\mathbf{S}_n = \left[ \begin{array}{ccc} 0 & -e_z & e_y \\ e_z & 0 & -e_x \\ -e_y & e_x & 0 \end{array} \right].
\end{equation}

\begin{figure}[H]
	\centering
	\includegraphics[width=0.49\textwidth]{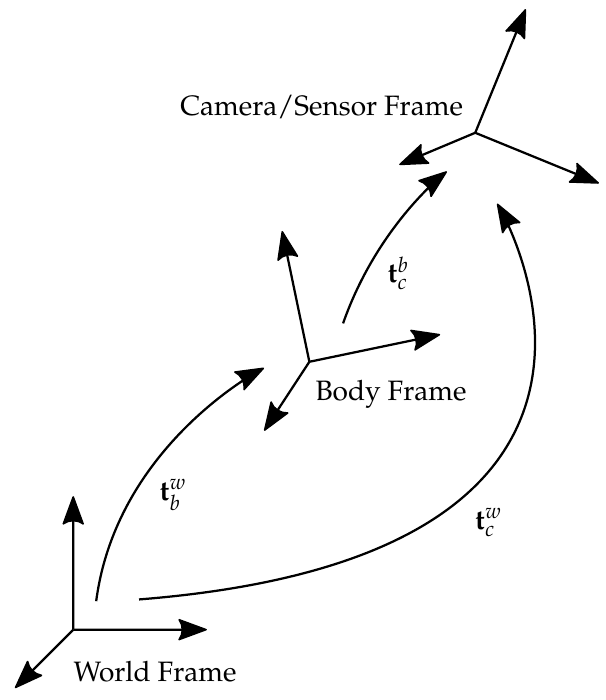}
	\caption{Summary of transforms referenced in this paper.}
	\label{fig:transforms}
\end{figure}

A complete 6 \ac{DOF} pose transform can be compactly represented with the three translation and three orientation parameters:

\begin{equation}
\mathbf{t} = [r_{x}, r_{y}, r_{z}, \phi_{x}, \phi_{y}, \phi_{z}]^T,
\end{equation}
or

\begin{equation}
\mathbf{t} = [r_{x}, r_{y}, r_{z}, \theta e_x, \theta e_y, \theta e_z]^T,
\end{equation}
depending on whether Euler or axis-angle conventions are used. The pose transforms of importance in this paper are the world to platform body transform $\mathbf{t}_b^w$, platform body to camera transform $\mathbf{t}_c^b$, and, combining these, the world to camera transform $\mathbf{t}_c^w$  (see Figure \ref{fig:transforms}). Note the sub- and superscripts: e.g.,~$\mathbf{t}_c^b$ denotes the translation and rotation of the camera axes with respect to the platform body.

By splitting the world pose of the camera $\mathbf{t}_c^w$ into a combination of the body pose $\mathbf{t}_b^w$ and the camera relative pose $\mathbf{t}_c^b$, $\mathbf{P}$ from Equation (\ref{eq:P}) can be shown as a function of the camera rotation and translation with respect to the body frame $\mathbf{R}_c^b$ and $\mathbf{p}_c^b$, and platform body rotation and translation with respect to the world frame $\mathbf{R}_b^w$ and $\mathbf{p}_b^w$:

\begin{equation}
\mathbf{P} = \mathbf{K} \left( {\mathbf{R}_c^b}^{-1} [\mathbf{I}_{3 \times 3} |-\mathbf{p}_c^b ] \right) 
\left(
\mleft[
\begin{array}{c|c}
{\mathbf{R}_b^w}^{-1} & \mathbf{0} \\
\hline
\mathbf{0}  & 1
\end{array}
\mright]
\mleft[
\begin{array}{c|c}
\mathbf{I}_{3 \times 3} & -\mathbf{p}_b^w \\
\hline
\mathbf{0} & 1
\end{array}
\mright]
\right)
\end{equation}

In our case,  $\mathbf{R}_b^w$ and $\mathbf{p}_b^w$ are provided by the navigation system, and $\mathbf{R}_c^b$ and $\mathbf{p}_c^b$ are the relative camera pose parameters we would like to estimate.

\subsection{Estimation of Calibration Pattern Points} \label{sec:cal_pattern_point_estimation}

\begin{figure}[H]
	\centering
	\includegraphics[width=0.49\textwidth]{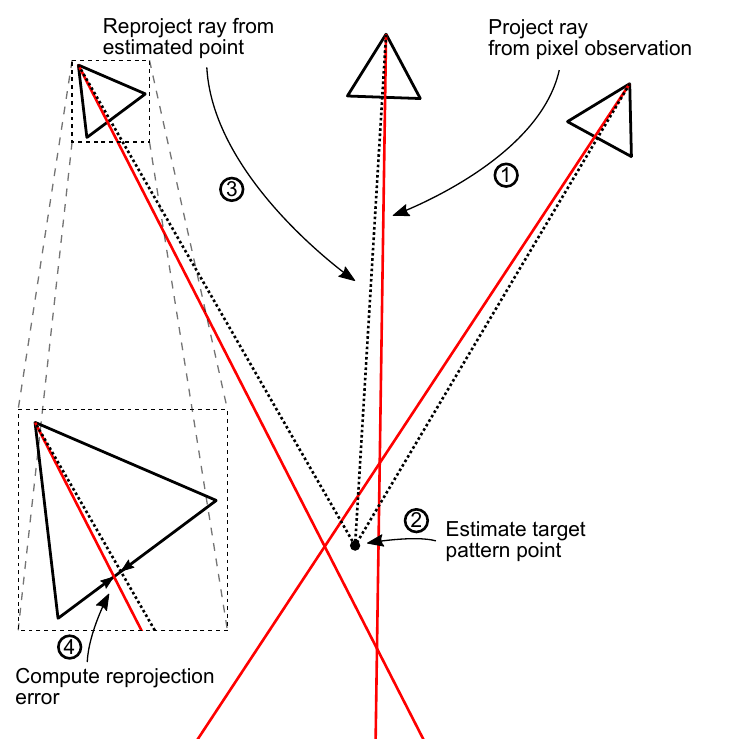} 
	\caption{Method summary. Rays corresponding to individual calibration pattern point observations are determined from pixel observations and camera poses. Calibration pattern point locations are then triangulated from all rays, and subsequently reprojected to the camera sensor. A reprojection error can then be computed by calculating the difference between the reprojected point and the pixel observation. Uncertainties are propagated through at each step, which facilitates the calculation of the uncertainty for the reprojection error, and subsequently a likelihood value, which is maximised by the optimiser.}
	\label{fig:method_summary}
\end{figure}

The first step of the proposed method involves estimating the location of calibration pattern points in world coordinates, as these are unknown and must be computed from the data. As shown in Figure~\ref{fig:method_summary}, rays are calculated for each pixel observation of each calibration pattern point, given the concurrent navigation system solution and camera pose proposal. Average point locations are determined by triangulating all rays corresponding to the same calibration pattern point. Uncertainties~for all inputs (pixel locations, navigation solutions and intrinsics) in the form of covariance matrices are propagated using the Jacobian of the point calculation function, yielding an uncertainty estimate (covariance matrix) for each calibration point estimate.

The proposed method starts with repeated imaging of points that can be uniquely identified. The use of a regular calibration pattern ensures points can be easily distinguished and is therefore recommended. The location of each pattern point $\mathbf{p}^w_k$ for $k \in \{1, 2, ..., M\}$ is estimated from all of its observation rays $i \in \{1, 2, ..., N\}$. There are $M$ points on the calibration pattern and the whole pattern is viewed $N$ times. For each point, we calculate the nearest points between all pairs of observation rays $(i,j)$ and apply a weighted average. Nearest points between rays are calculated as follows \citep{vnr_concise_math}:

\begin{equation} \label{eq:line_closest_point}
\begin{split}
\mathbf{p}^w_{k,ij} = &\mathbf{p}^w_{c,k,i} + \\
&\frac{(\mathbf{p}^w_{c,k,j} - \mathbf{p}^w_{c,k,i}) \cdot \mathbf{n}_{k,ij}}
{( \mathbf{p}^w_{s,k,i} - \mathbf{p}^w_{c,k,i} ) \cdot \mathbf{n}_{k,ij}}
( \mathbf{p}^w_{s,k,i} - \mathbf{p}^w_{c,k,i} ),
\end{split}
\end{equation}
where

\begin{equation}
\begin{split}
\mathbf{n}_{k,ij} = &( \mathbf{p}^w_{s,k,j} - \mathbf{p}^w_{c,k,j} ) \times \\
&[ ( \mathbf{p}^w_{s,k,i} - \mathbf{p}^w_{c,k,i} ) \times
( \mathbf{p}^w_{s,k,j} - \mathbf{p}^w_{c,k,j} ) ].
\end{split}
\end{equation}

We could estimate $\mathbf{p}^w_k$ as the unweighted mean of all $\mathbf{p}^w_{k,ij}$ for a given pattern point $k$, but some estimates are more certain than others given the conditions of how they were measured. A more accurate estimate is obtained using a weighted average according to the uncertainty.  The uncertainty of each point $\mathbf{p}^w_{k,ij}$ can be obtained by computing its Jacobian $\mathbf{J}_{\mathbf{p}^w_{k,ij}}$ with respect to all input values. Also required are the uncertainties of the pixel and platform pose observations for each ray, expressed as covariance matrices, $\mathbf{Q}_{uv,k,i}$, $\mathbf{Q}_{\mathbf{t}_{b,k,i}^w}$, $\mathbf{Q}_{uv,k,j}$ and $\mathbf{Q}_{\mathbf{t}_{b,k,j}^w}$, as well as intrinsic and extrinsic parameter covariances, $\mathbf{Q}_{int}$ and $\mathbf{Q}_{\mathbf{t}_{c}^b}$. Although line scan cameras have only one pixel coordinate ($u$), there is also uncertainty in the second coordinate $v$, because a point elicits a pixel response if it is located within the camera's \ac{IFOV}, not necessarily directly on the scan line. $\mathbf{Q}_{int}$ and $\mathbf{Q}_{\mathbf{t}_{c}^b}$ contain variances and covariances of the intrinsic camera parameters  and the relative camera pose respectively. All the input covariance matrices are combined into one matrix $\mathbf{Q}_{k,ij}$:

\begin{equation} \label{eq:p_input_cov}
\arraycolsep=0.0pt\def\arraystretch{1.0}
\mathbf{Q}_{k,ij} = 
\left[ 
\begin{array}{cccccc}
\mathbf{Q}_{uv,k,i} & \mathbf{0} & \mathbf{0} & \mathbf{0} & \mathbf{0} & \mathbf{0} \\
\mathbf{0} & \mathbf{Q}_{\mathbf{t}_{b,k,i}^w} & \mathbf{0} & \mathbf{0} & \mathbf{0} & \mathbf{0} \\
\mathbf{0} & \mathbf{0} & \mathbf{Q}_{uv,k,j} & \mathbf{0} & \mathbf{0} & \mathbf{0} \\
\mathbf{0} & \mathbf{0} & \mathbf{0} & \mathbf{Q}_{\mathbf{t}_{b,k,j}^w} & \mathbf{0} & \mathbf{0} \\
\mathbf{0} & \mathbf{0} & \mathbf{0} & \mathbf{0} & \mathbf{Q}_{int} & \mathbf{0} \\
\mathbf{0} & \mathbf{0} & \mathbf{0} & \mathbf{0} & \mathbf{0} & \mathbf{Q}_{\mathbf{t}_{c}^b}
\end{array}
\right].
\end{equation}

No correlation between the navigation solutions of the two rays is assumed, which is reasonable if the two observations are sufficiently separated in time. $\mathbf{Q}_{k,ij}$ and $\mathbf{J}_{\mathbf{p}^w_{k,ij}}$ can now be used to compute the uncertainty of $\mathbf{p}^w_{k,ij}$ (Equation (\ref{eq:line_closest_point})) as covariance matrix $\mathbf{\Sigma}_{\mathbf{p}^w_{k,ij}}$:

\begin{equation}
\mathbf{\Sigma}_{\mathbf{p}^w_{k,ij}} = \mathbf{J}_{\mathbf{p}^w_{k,ij}} \mathbf{Q}_{k,ij} \mathbf{J}_{\mathbf{p}^w_{k,ij}}^T.
\end{equation}

Because we wish to estimate both $\mathbf{t}_{c}^b$ and $\mathbf{Q}_{\mathbf{t}_{c}^b}$ with respect to all error sources other than the camera pose, we set all elements of the $6 \times 6$ covariance matrix $\mathbf{Q}_{\mathbf{t}_{c}^b}$ to zero temporarily \citep{underwood2009reliable,underwood2010error}. Each point $\mathbf{p}^w_k$ on the calibration pattern can then be estimated by computing an average that is weighted according to the covariances \citep{james2006statistical}:

\begin{equation}
\mathbf{W}_{k,ij} = \mathbf{\Sigma}_{\mathbf{p}^w_{k,ij}}^{-1},
\end{equation}

\begin{equation}
\mathbf{\Sigma}_{\mathbf{\hat{p}}^w_{k}} = \left(\sum_{i}^{N} \sum_{j}^{N} \mathbf{W}_{k,ij} \right)^{-1},
\end{equation}

\begin{equation} \label{eq:avg_points}
\mathbf{\hat{p}}^w_{k} = \mathbf{\Sigma}_{\mathbf{\hat{p}}^w_{k}} \left(\sum_{i}^{N} \sum_{j}^{N} \mathbf{W}_{k,ij} \mathbf{p}^w_{k,ij} \right).
\end{equation}

This ensures that the contribution of each closest point for each ray pair ($\mathbf{p}^w_{k,ij}$) is weighted according to its certainty, taking into account navigation system uncertainty or challenging viewpoint geometry (such as a small angle between the two rays).

\subsection{Calculation of Reprojection Error and Likelihood Optimisation} \label{sec:reproj_error_and_likelihood}

Once estimates and uncertainties of each calibration pattern point have been obtained, they are reprojected to the camera for each observation, which allows calculating an error between each of the observed pixel locations and the reprojected pixels (see Figure \ref{fig:method_summary}). The uncertainties of all inputs and calibration pattern point estimates are also propagated through, which yields an uncertainty value for each reprojection error. This enables the calculation of an overall likelihood value of the data given a camera pose proposal. The optimiser maximises this likelihood by varying the camera pose to arrive at an estimate.

For each observation $i$, $\mathbf{\hat{p}}^w_{k}$ can be reprojected according to Equation (\ref{sec:pinhole_projection}), given a $\mathbf{t}_{c}^b$ and corresponding navigation system solution $\mathbf{t}_{b,k,i}^w$. The reprojection error is calculated as the Euclidean distance between the reprojected and observed pixel locations: 

\begin{equation} \label{eq:reproj_error}
e_{k,i} = \sqrt{(u_{k,i} - \hat{u}_{k,i})^2 + (v_{k,i} - \hat{v}_{k,i})^2}.
\end{equation}

The reprojection is two dimensional, because non-optimal $\mathbf{t}_{c}^b$ can result in reprojected pixels that deviate from the one dimensional scan line ($\hat{v}_{k,i} \neq 0$), but $v_{k,i}$ is assumed to be 0. The variance of the reprojection error can also be computed using the input covariance matrix and Jacobian $\mathbf{j}_{e_{k,i}}$:

\begin{equation} \label{eq:reproj_error_var}
\sigma_{e_{k,i}}^2 = \mathbf{j}_{e_{k,i}}^T \mathbf{Q}_{k,i} \mathbf{j}_{e_{k,i}},
\end{equation}
where

\begin{equation}
\mathbf{Q}_{k,i} = 
\left[ 
\begin{array}{ccccc}
\mathbf{\Sigma}_{\mathbf{\hat{p}}^w_{k,l}} & \mathbf{0} & \mathbf{0} & \mathbf{0} & \mathbf{0} \\
\mathbf{0} & \mathbf{Q}_{uv,k,i} & \mathbf{0} & \mathbf{0} & \mathbf{0} \\
\mathbf{0} & \mathbf{0} & \mathbf{Q}_{\mathbf{t}_{b,k,i}^w} & \mathbf{0} & \mathbf{0} \\
\mathbf{0} & \mathbf{0} & \mathbf{0} & \mathbf{Q}_{int} & \mathbf{0} \\
\mathbf{0} & \mathbf{0} & \mathbf{0} & \mathbf{0} & \mathbf{Q}_{\mathbf{t}_{c}^b} 
\end{array}
\right].
\end{equation}

The Jacobian $\mathbf{j}_{e_{k,i}}$ is lower case because it is only one dimensional in this instance, since $e_{k,i}$ is a scalar value. As in Equation (\ref{eq:p_input_cov}), we again set all elements of $\mathbf{Q}_{\mathbf{t}_{c}^b}$ to zero. The log likelihood of a transform $\mathbf{t}_{c}^b$ given the observations can then be estimated as,

\begin{equation}
log\Lambda = - \sum_{k}^{M} \sum_{i}^{N} \frac{e_{k,i}^2} {2\sigma_{e_{k,i}}^2}.
\label{eq:log_likelihood}
\end{equation}

The objective is to maximise $log\Lambda$, by varying the 6-\ac{DOF} $\mathbf{t}_c^b$ vector. This can be achieved using standard optimisation methods to minimise the negative log likelihood:

\begin{equation}
\argmin{\mathbf{t}_{c}^{b}} -log\Lambda = \argmin{\mathbf{t}_{c}^{b}} \sum_{k}^{M} \sum_{i}^{N} \frac{e_{k,i}^2} {2\sigma_{e_{k,i}}^2},
\label{eq:neg_log_likelihood}
\end{equation}

$log\Lambda$ is fully recalculated at each optimisation iteration, which includes the triangulation of calibration pattern points and calculation of their reprojection error.

\subsection{Variance-Covariance Matrix Estimation} \label{sec:uncertainty_estimation}
Once the relative camera pose $\mathbf{t}_{c}^{b}$ has been determined, it is desirable to approximate the covariance matrix of the solution, which provides an estimate of how uncertain the six relative camera pose parameters are. In combination with covariances of other parameters, such as the navigation system solution, this also allows mapping accuracy to be quantified. In other words, the result provides values for $\mathbf{Q}_{\mathbf{t}_c^b}$, completing the full covariance matrix (see Equation (\ref{eq:proj_cov_matrix}) in Section \ref{sec:method_projection}). Note that all elements \textls[-15]{of $\mathbf{Q}_{\mathbf{t}_c^b}$ are set to zero for its estimation and optimisation, as previously mentioned in Sections \ref{sec:cal_pattern_point_estimation} and \ref{sec:reproj_error_and_likelihood}.}

The proposed approach is based on similar work done with lidar sensors \citep{underwood2009reliable,underwood2010error}, but the details differ because 1D cameras do not directly provide depth information. We propose a random sampling based method, where a set of sample sensor to body transforms are selected using a \ac{MCMC} algorithm~\citep{foreman2013emcee}, which differs from the \ac{MC} importance sampling approach in \cite{underwood2009reliable,underwood2010error}. This provides greater sampling efficiency and avoids the need to manually define a sampling region. The algorithm is guided by the likelihood of each relative camera pose sample, which governs the selection of the next sample.

There are several \ac{MCMC} variations, but they all share the property that each sample is selected based on the previous. For a large number of samples, the distribution tends towards the probability distribution that is being sampled from (i.e., $\Lambda$ in this paper) \citep{MacKay2008}. For further details about \ac{MCMC} sampling, the reader is referred to the numerous resources available on the topic \citep{MacKay2008,foreman2013emcee}. The \ac{MCMC} algorithm provides a list of samples $\{t_{c,1}^b,t_{c,2}^b, ..., t_{c,l}^b, ..., t_{c,r}^b\}$, which are distributed according to $\Lambda$, from which the covariance can be computed as:

\begin{equation} \label{eq:camera_pose_covariance}
\mathbf{Q}_{\mathbf{t}_c^{b}\textbf{*}} = \frac{1}{r-1} \mathlarger{\sum}_{l=1}^{r} \left(\mathbf{t}_{c,l}^b-\mathbf{\bar{t}}_{c}^b \right) \left(\mathbf{t}_{c,l}^b-\mathbf{\bar{t}}_{c}^b \right)^T,
\end{equation}
where

\begin{equation}
\mathbf{\bar{t}}_c^b = \frac{1}{r} \sum_{l=1}^{r} \mathbf{t}_{c,l}^b,
\end{equation}

\section{Materials and Methods} \label{sec:methods}
This section outlines the equipment and methods used to obtain the data and analyse the results. A planar calibration pattern was placed in the environment and imaged from several different orientations using a line scanning camera mounted to two different ground based robotic platforms. A navigation system mounted to each platform recorded the 6 \ac{DOF} position and orientation of the platforms ($\mathbf{t}_b^w =~[r_{b,x}^w, r_{b,y}^w, r_{b,z}^w, \phi_{b,x}^w, \phi_{b,y}^w, \phi_{b,z}^w]$) throughout the acquisition period. Image pixel locations of calibration pattern points and matching robot poses were then used to estimate the relative camera pose using an iterative optimisation algorithm. Finally, the uncertainty of the camera pose estimate in the form of a covariance matrix was approximated using \ac{MCMC}.

First the ground based mobile platforms and associated sensors used to acquire data are introduced, followed by a description of the data acquisition process and extraction of pattern point observations. The implementation of the method presented in Section \ref{sec:general_approach} is outlined, which includes the optimisation and an outlier removal process. Methods for mapping image data and comparing camera poses are presented, as required for the analysis of the results, and a method is presented to calculate the basin of attraction, to assess the sensitivity of the process to the initial camera pose.

\subsection{Mobile Sensing Platforms} \label{sec3.1}
\begin{figure}[H]
	\centering
	\begin{subfigure}{.49\textwidth}
		\includegraphics[width=1.\textwidth]{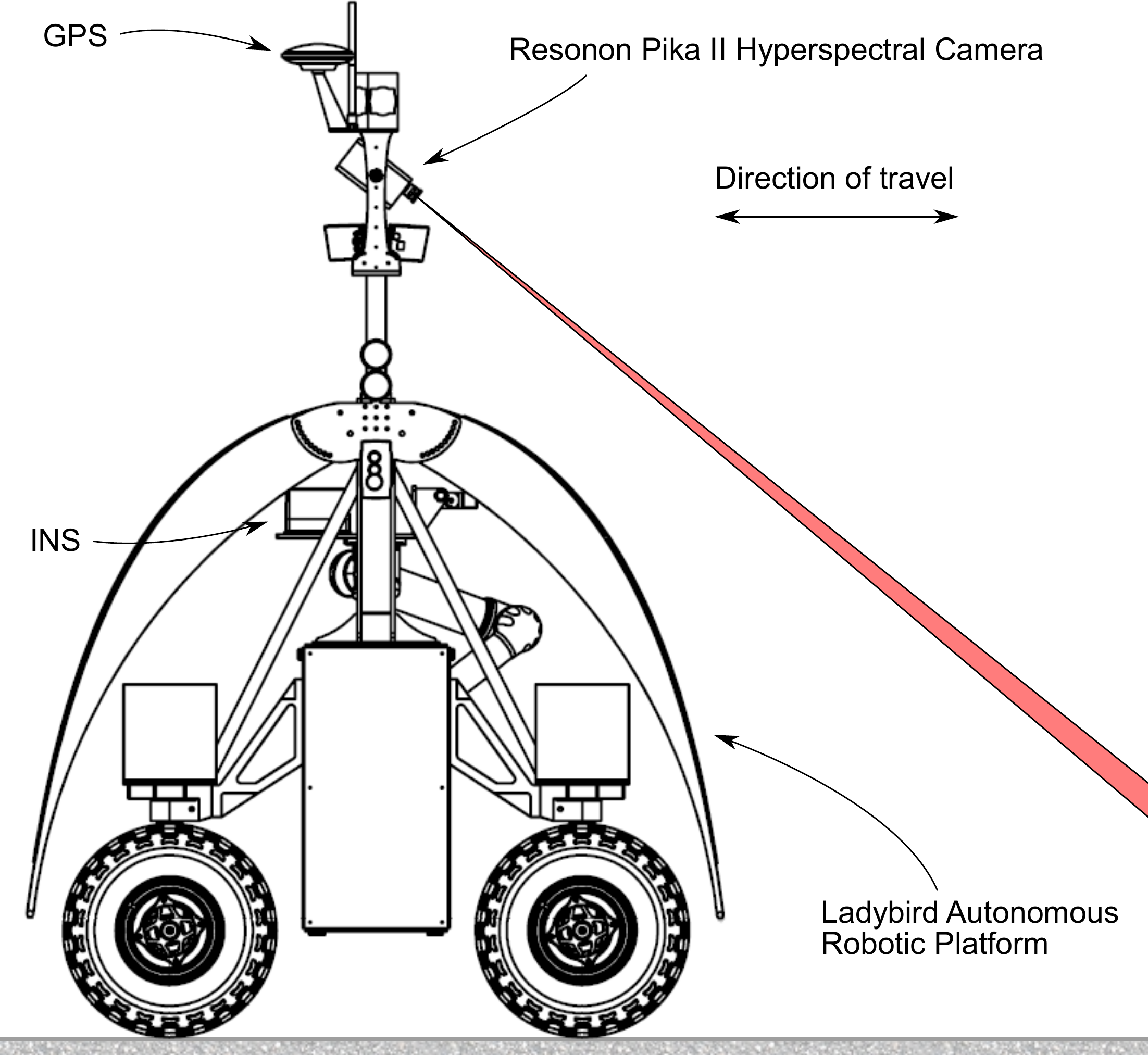} 
		\caption{Ladybird robotic platform}
		\label{fig:ladybird_platform}
	\end{subfigure} 
	\begin{subfigure}{.49\textwidth}
		\includegraphics[width=1.\textwidth]{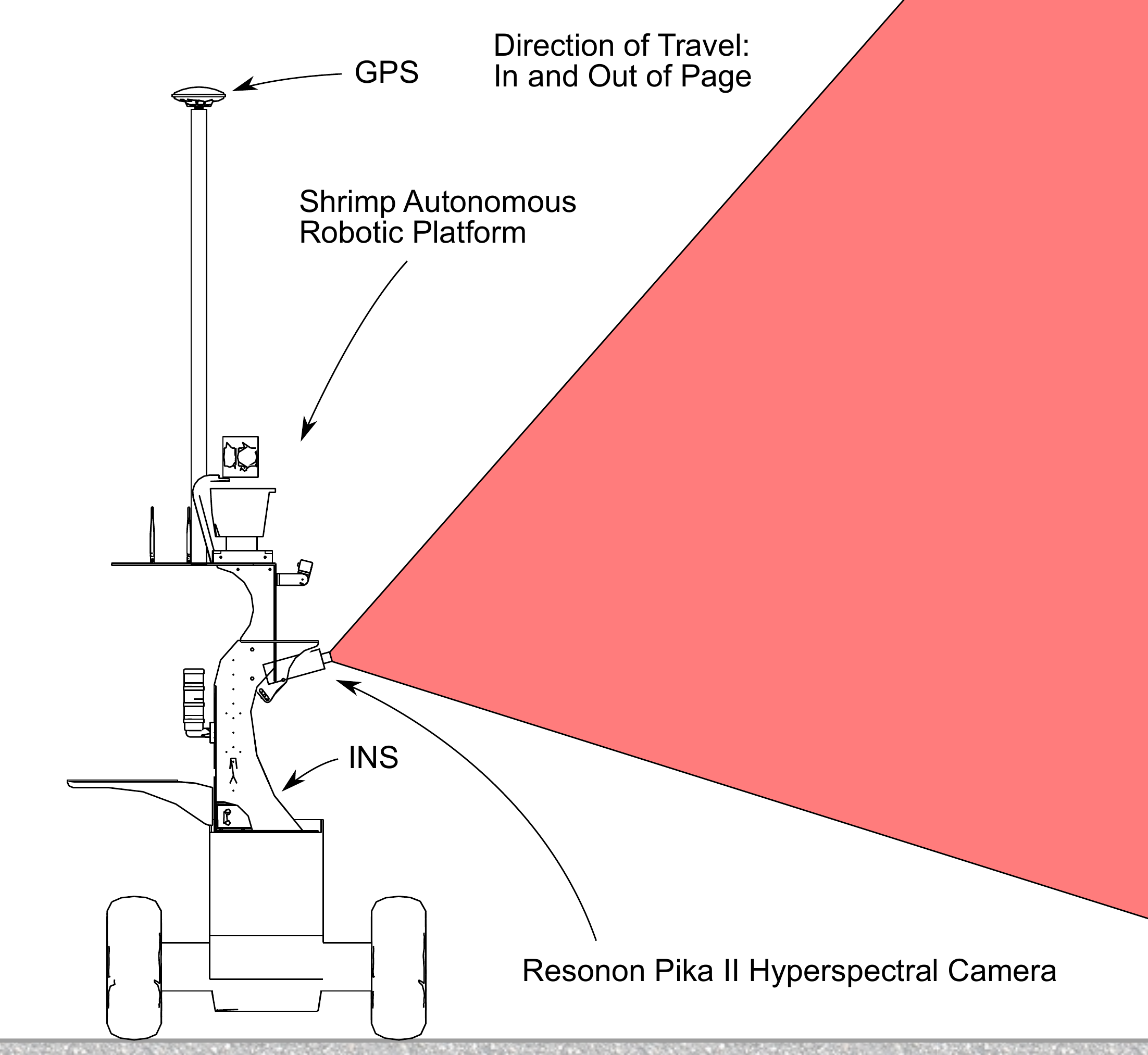} 
		\caption{Shrimp robotic platform}
		\label{fig:shrimp_platform}
	\end{subfigure}
	\vspace{-12pt}
	\caption{The Ladybird (\textbf{a}) and Shrimp (\textbf{b}) robotic platforms and sensor configurations.}
	\label{fig:platforms}
\end{figure}

A line scanning hyperspectral camera was mounted to two different robotic platforms, Ladybird and Shrimp (Figure \ref{fig:platforms}). \textls[-15]{Both were designed and built at the \ac{ACFR} at The University of Sydney as flexible tools to support a range of research applications \citep{underwoodplantphenomics,underwoodreal,wendel2017illumination,stein2016image,bargoti2017image,underwood2016mapping}. }The sensor suite on both platforms includes a \ac{RTK}/\ac{GPS}/\ac{INS}, which provides platform pose and covariance estimates (details in Table \ref{tab:configuration}). The \ac{GPS} units on both platforms are identical, but the Shrimp platform uses a lower grade \ac{IMU} than the Ladybird platform.

Line scan image data were acquired with a Resonon Pika II \ac{VNIR} line scanning camera that was mounted to the Ladybird and Shrimp robots in a push broom configuration. For the Ladybird, the camera was oriented such that the scan line is horizontal, pitched down for scanning the ground surface (Figure \ref{fig:ladybird_platform}). On Shrimp, the camera was mounted such that the scan line is vertical, and pitched upwards slightly to allow scanning of upright objects (Figure \ref{fig:shrimp_platform}). The camera produces hyperspectral images of 648 spatial by 244 spectral pixels (spectral resolution of ~2 nm from 390.9--887.4 nm) at a rate of 133 frames per second and native bit depth of 12. For the purposes of this paper, the spectral dimension was averaged to produce 648 pixel monochrome scan lines. Apart from this averaging step, the method described in this paper is not particular to hyperspectral cameras and may be applied equally to other types of line scanning imagers. Schneider Cinegon 6 mm and 8~mm objective lenses were used for Shrimp and Ladybird respectively, and manually focused with a checker board at the typical distance to the scene. The principal point of the camera/lens combination was assumed to be at the centre of the line scan ($u_0=323$), the focal length was assumed to be as per the manufacturer supplied measurements (see Table \ref{tab:configuration}), and distortion was assumed to be zero. Hand~measured pose estimates and manufacturer supplied lens details are shown in Table  \ref{tab:configuration}.

\unskip
\newcommand*{\tabindent}{\hspace*{0.25cm}}
\begin{table}[H]
	\centering
	\tablesize{\footnotesize}
	\begin{tabulary}{0.49\textwidth}{lcc}
	\toprule
		& \textbf{Ladybird} & \textbf{Shrimp} \\
		\hline \noalign{\vskip 1mm} 
		\multicolumn{3}{l}{\textbf{Manually Measured Camera Pose} \boldmath{$t_c^b$}} \\
		\midrule
		\tabindent $r_{c,x}^b, r_{c,y}^b,r_{c,z}^b$ (m) & $0.2,0.0,-0.8$ & $0.0,-0.2,-0.5$ \\
\tabindent$\phi_{c,x}^b, \phi_{c,y}^b, \phi_{c,z}^b$ (\degree) & $-56.0,0.0,-90.0$ & $0.0,105.0,-90.0$ \\
\midrule
		\multicolumn{3}{l}{\textbf{Camera Lens Details}} \\
		\midrule
		\tabindent Manufacturer & Schneider & Schneider \\
		\tabindent Model & Cinegon 8 mm & Cinegon 6 mm \\
		\tabindent Focal length & 8.2 mm & 6.2 mm \\
		\tabindent Approx. aperture & f/2.5 & f/3.0 \\
		\tabindent IFOV & 1.88 mrad & 2.5 mrad \\
		\midrule
		\multicolumn{3}{l}{\textbf{Navigation System Details}} \\
		\midrule
		\tabindent Manufacturer & Novatel & Novatel \\
		\tabindent GPS receiver & ProPak-G2plus & ProPak-G2plus \\
		\tabindent IMU & Honeywell HG1700 & IMU-CPT\\
\bottomrule
	\end{tabulary}
	\caption{Platform configurations.}
	\label{tab:configuration}
\end{table}

Initial pose estimates were measured by hand with the mobile platforms on a level surface using measuring tape for translational offsets, and a digital inclinometer (SPI Pro 3600) for angular offsets around the robots' horizontal $x$ and $y$ axes. Angular offsets around the robots' vertical $z$ axis were assumed to be the intended mounting orientations, which are in increments of 90\degree \,for both platforms. Note that if the camera is mounted at angles that are clearly not in 90\degree \,increments, referring to a CAD model is recommended. Hand measured translation parameters ($r_{c,x}^b$, $r_{c,y}^b$ and $r_{c,z}^b$) were assumed to have a standard deviation ($\sigma$) of 0.1 m and orientation parameters ($\phi_{c,x}^b$, $\phi_{c,y}^b$ and $\phi_{c,z}^b$) were assumed to have a $\sigma$ of 2\degree. 

\subsection{Data Acquisition} \label{sec:data_aquisition}
\begin{figure}[H]
	\centering
	\begin{subfigure}{.49\textwidth}
		\includegraphics[width=1.\textwidth]{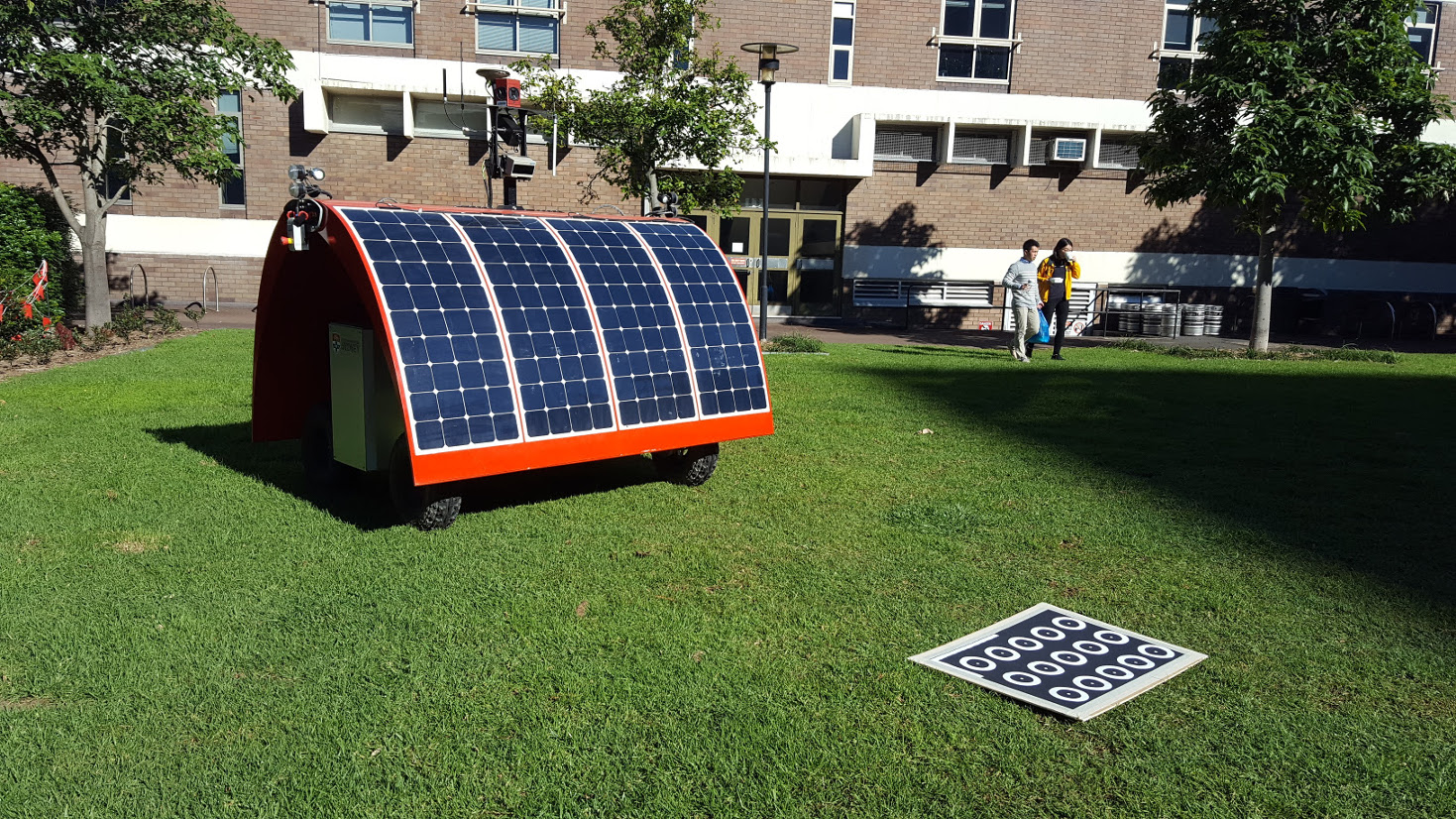} 
		\caption{Ladybird data acquisition}
		\label{fig:ladybird_setup}
	\end{subfigure} 
	\begin{subfigure}{.49\textwidth}
		\includegraphics[width=1.\textwidth]{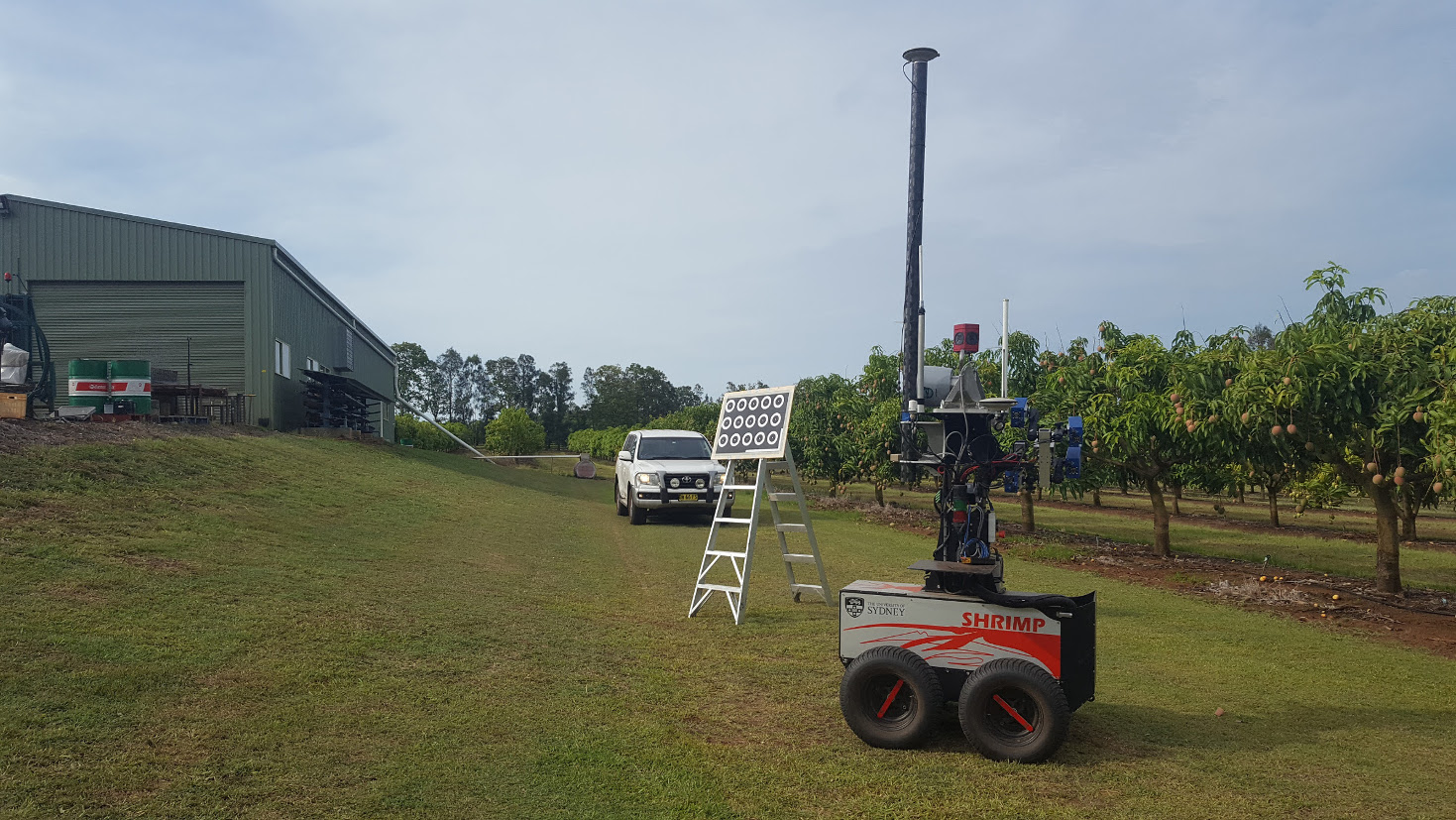} 
		\caption{Shrimp data acquisition}
		\label{fig:shrimp_setup}
	\end{subfigure}
	\begin{subfigure}{.4\textwidth}
		\includegraphics[width=1.\textwidth]{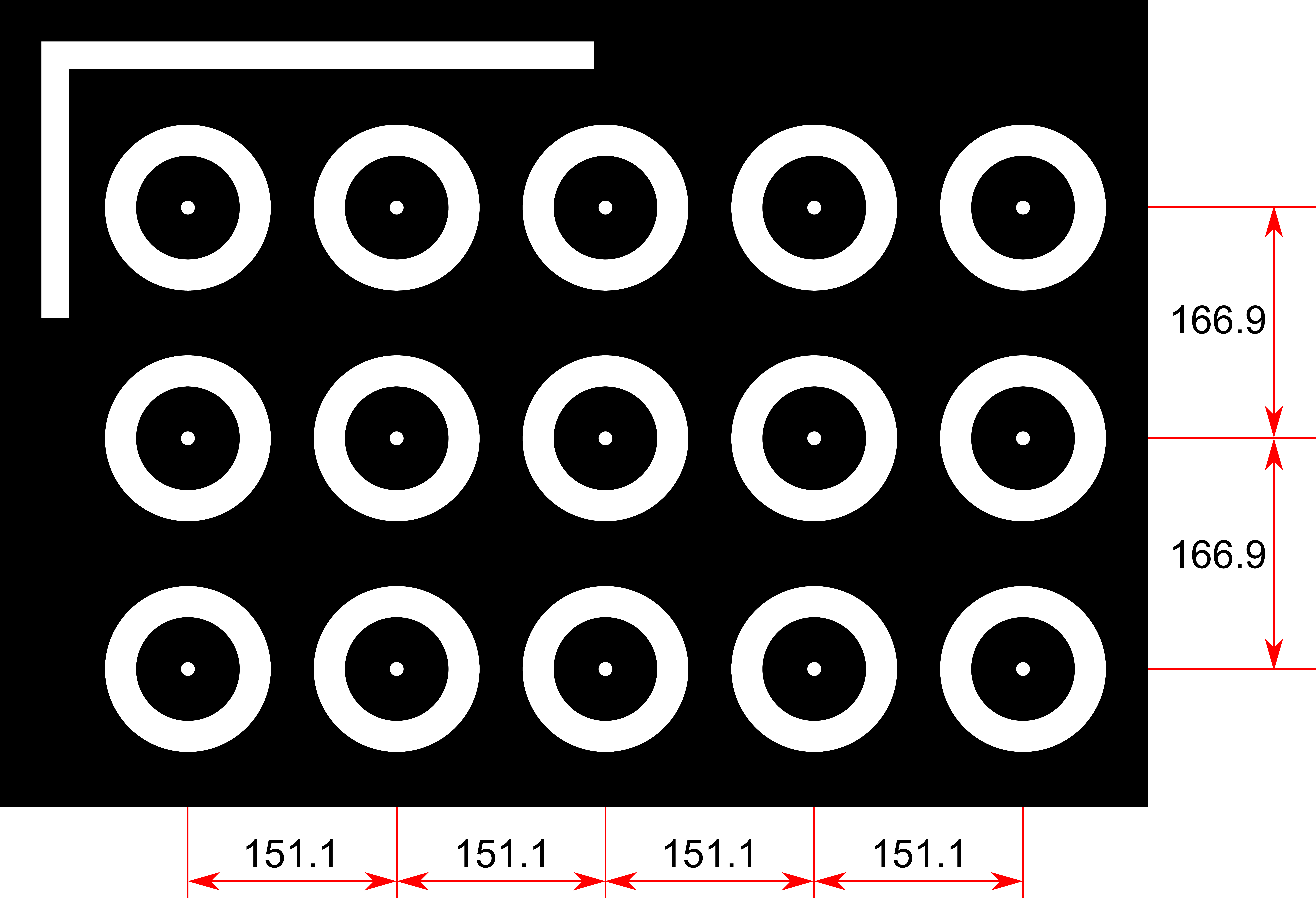} 
		\caption{Calibration pattern}
		\label{fig:calibration_pattern}
	\end{subfigure}
	\vspace{-12pt}
	\caption{Data acquisition configuration and location for Ladybird (\textbf{a}) and Shrimp (\textbf{b}). The calibration pattern used in both instances is shown in (\textbf{c}). Dimensions are in mm. The centre dots are 10 mm in diameter, sized to be as small as possible while still being visible in the image data in order to maximise labelling accuracy. The outer rings help with locating the points in the data for labelling, and are 75--120~mm, inner to outer diameter. Note that these dimensions do not need to be known for the optimisation procedure, nor is it necessary to print them in any particular strict arrangement, though a regular pattern is recommended to allow easy identification of each point.}
	\label{fig:setup}
\end{figure}

A calibration pattern with 15 points arranged in a 3 $\times$ 5 pattern was printed to an A1 size sheet of paper and mounted to a flat rigid plywood board (see Figure \ref{fig:calibration_pattern}). The pattern was designed to maximise contrast for efficient extraction of pattern points. A corner shape was added to one side of the pattern to facilitate unique identification of each point. It is not necessary to know the pattern's dimensions for recovery of the platform to camera pose, as each point is treated independently during the calibration. This also means that theoretically a single point with sufficient observations could be used for calibration. However, we added more pattern points since there is no significant practical cost, efficiently increasing the amount of data obtained.

\begin{figure}[H]
	\centering
	\begin{subfigure}{.49\textwidth}
		\includegraphics[width=1.\textwidth]{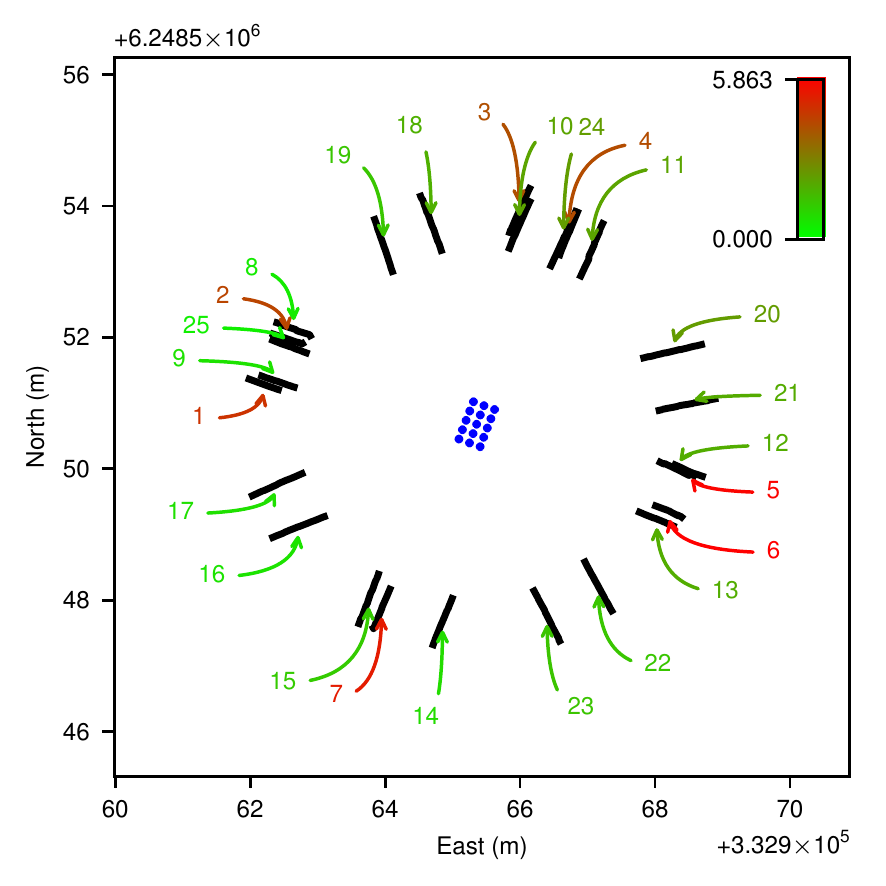} 
		\caption{Ladybird}
		\label{fig:ladybird_runs}
	\end{subfigure}
	\begin{subfigure}{.49\textwidth}
		\includegraphics[width=1.\textwidth]{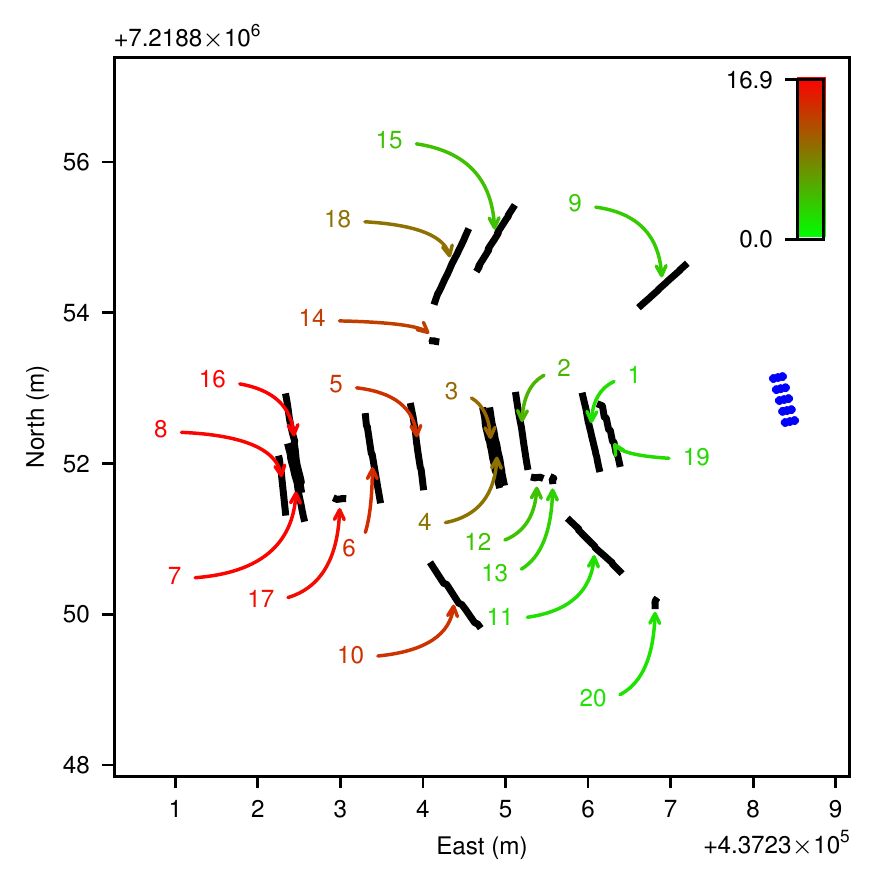}
		\caption{Shrimp}
		\label{fig:shrimp_runs}
	\end{subfigure}
	\vspace{-12pt}
	\caption{Top down view of platform positions (from navigation system) during each observation of the calibration pattern for Ladybird (\textbf{a}) and Shrimp (\textbf{b}). Each observation provides a complete view of all points on the calibration pattern, as well as concurrent navigation system solutions. The~calibration pattern points are indicated with blue dots. The observation runs are numbered, and coloured according to the norm of vehicle body pitch and roll (i.e., $\sqrt{{\phi_{b,x}^w}^2 + {\phi_{b,y}^w}^2}$), as indicated by the colour bars in degrees. Note that in (\textbf{b}) the calibration pattern was upright (mounted to a ladder), which is why the points appear more closely clustered from the top-down perspective. Some of the observation runs appear very short in (\textbf{b}). This is because in these instances the platform scanned the pattern by rotating on the spot.}
	\label{fig:all_runs}
\end{figure}

For the Ladybird platform, the pattern board was placed on relatively flat ground (see Figure \ref{fig:ladybird_setup}). As shown in Figure \ref{fig:all_runs}, the pattern was scanned from several directions around a circle with the calibration pattern in the centre. Two types of scans were performed, one with the robot's wheels flat on the ground and one with one side of the robot elevated by driving over an aluminium channel. This raised two of the wheels by approximately 100~mm, inducing a roll of approximately 4\degree{}. For the Shrimp platform, the same calibration pattern was mounted to a ladder in an approximately vertical orientation (see Figure \ref{fig:shrimp_setup}). In this case data were acquired next to a hill with various orientations and positions with respect the pattern, where the hill caused continuously variable roll and pitch, up to approx. 17 \degree  \,(see Figure \ref{fig:shrimp_runs}). For both platforms, body orientation was intentionally varied as much as possible in an attempt to maximise observability of parameters \citep{underwood2009reliable}. The robots were manually operated throughout the acquisition period, and care was taken to move slowly and smoothly while the calibration pattern was imaged.

All data was timestamped allowing association between individual scan lines and platform pose solutions. Localisation uncertainties reported by the navigation system are shown in Table \ref{tab:median_stds} as median standard deviations (i.e., square root of the diagonals of the covariance matrices only) for the acquisition runs, which illustrates that the navigation system in the Ladybird platform is able to provide body pose estimates with much greater certainty than the navigation system on Shrimp, due to the higher grade \ac{IMU}.

\begin{table}[H]
	\centering
	\tablesize{\footnotesize}
	\caption{Median navigation system uncertainties as 1 standard deviation.}	
	\tabcolsep=0.11cm
	\begin{tabulary}{0.49\textwidth}{ccccccc}
	\toprule
		\textbf{Platform} & \boldmath{$\sigma_{r_{b,x}^w}$} \textbf{(m)} & \boldmath{$\sigma_{r_{b,y}^w}$} \textbf{(m)} & \boldmath{$\sigma_{r_{b,z}^w}$} \textbf{(m)} & \boldmath{$\sigma_{\phi_{b,x}^w}$} \textbf{(\degree)} & \boldmath{$\sigma_{\phi_{b,y}^w}$} \textbf{(\degree)} & \boldmath{$\sigma_{\phi_{b,z}^w}$} \textbf{(\degree)} \\
		\hline \noalign{\vskip 1mm} 
		Ladybird & \num{1.052e-02} & \num{1.305e-02} & \num{1.118e-02} & \num{2.362e-01} & \num{2.636e-01} & \num{1.053e-01} \\
		Shrimp & \num{4.520e-02} & \num{4.369e-02} & \num{4.887e-02} & \num{7.534e-01} & \num{7.284e-01} & \num{8.416e-01}\\
\bottomrule
	\end{tabulary}
	\label{tab:median_stds}
\end{table}
\unskip

\subsection{Pattern Pixel Extraction}
\begin{figure}[H]
	\centering
	\includegraphics[width=.9\textwidth]{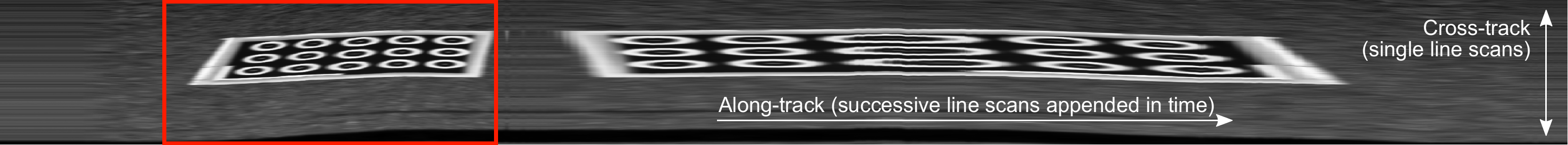}
	\caption{Example grey scale image obtained by appending successive line scans from Ladybird (without pose compensation). A single view of all points on the calibration pattern is referred to as an observation in this paper, including concurrent navigation system data. The red box indicates one such observation, shown as a more detailed close-up view in Figure \ref{fig:target_point_extraction}. The two views of the calibration pattern shown in this figure were obtained by driving the platform forward, producing the first view, and then backwards, giving the second view, which is therefore a mirrored version of the first. The~second view appears more stretched because it was scanned more slowly, generating more line scans, given the same fixed frame rate. Note that the observations shown in Figure \ref{fig:ladybird_runs} only include one observation from each forward-reverse pair, because they provide almost the same information (i.e.,~similar navigation system solutions).}
	\label{fig:sample_full_scan}
\end{figure}

\begin{figure}[H]
	\centering
	\includegraphics[width=.8\textwidth]{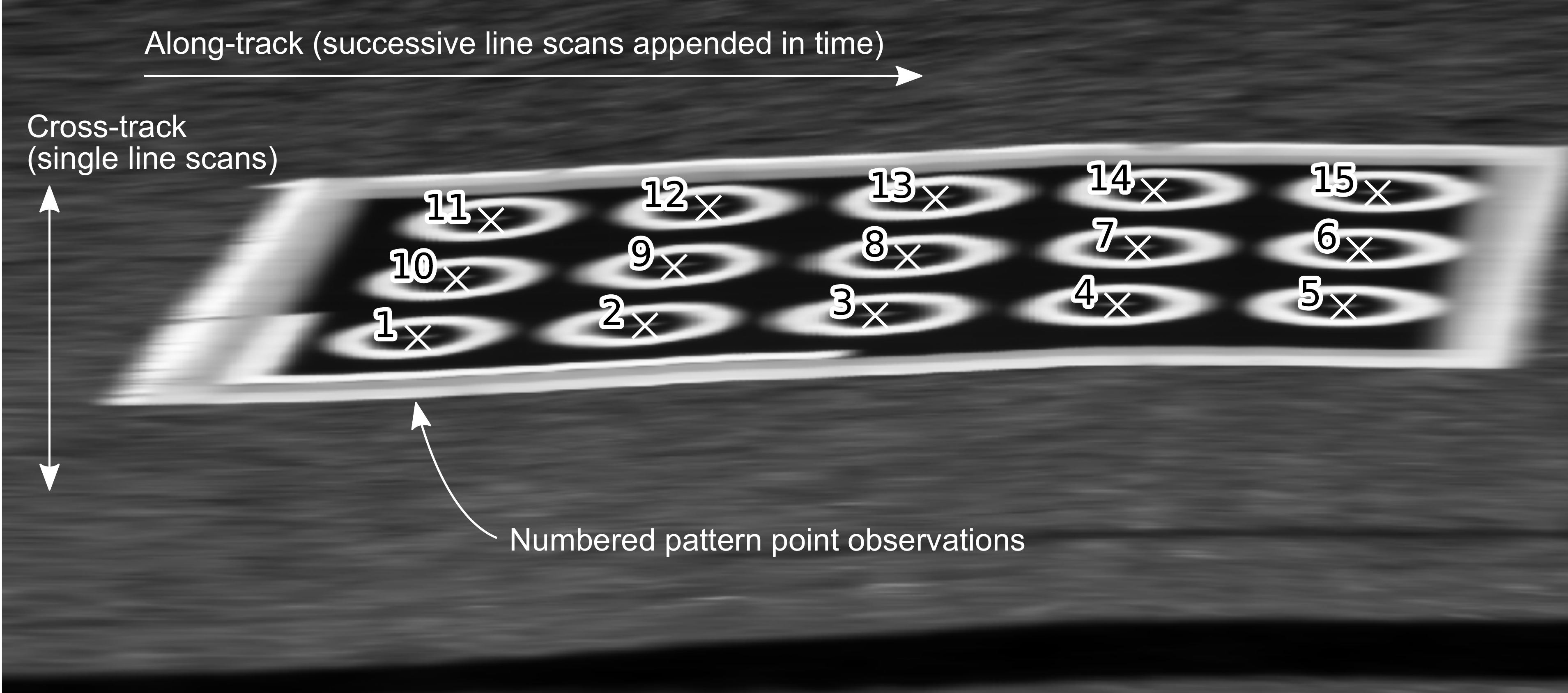}
	\caption{Close-up view of the observation indicated in Figure \ref{fig:sample_full_scan}, showing manually labelled calibration point locations (white crosses) with ID numbers. Each 648 pixel line scan is vertical and concatenated horizontally. A similar image was generated for each observation of the calibration pattern prior to manually labelling each point. Care was taken to ensure the numbering scheme remained consistent for all observations of the pattern.}
	\label{fig:target_point_extraction}
\end{figure}

Approximate pixel locations of points on the calibration pattern were selected manually by appending successive line scans to form a rectangular image and selecting individual pattern points in order (see Figures \ref{fig:sample_full_scan} and \ref{fig:target_point_extraction}). Note that line scans were concatenated naively, ignoring camera or body pose data (i.e., not mapped or georeferenced). This worked well for this purpose, because the platforms were moved slowly and smoothly while the calibration pattern was scanned. Particular~care was taken to ensure that point ID numbers were consistent for all observations of the calibration pattern. Pixel~locations were then refined to sub pixel precision by extracting a $10 \times 10$ patch around the selected points and resizing it to $100 \times 100$ pixels using bi-cubic interpolation. The intensity peak closest to the centre was taken as the pattern point pixel location. Along-track, the closest time stamp was used to obtain the corresponding navigation solution. This provides pixel position $u_{k,i}$ and \textls[-15]{platform pose $[r_{b,x}^w,r_{b,y}^w,r_{b,z}^w,\phi_{b,x}^w,\phi_{b,y}^w,\phi_{b,z}^w]^T$, which are necessary for calibration according to Equation~(\ref{eq:neg_log_likelihood}).}

\unskip

\subsection{Optimisation and Uncertainty Estimation} \label{sec:optimisation_method}
Optimisation was performed using the Powell optimiser algorithm provided by the SciPy python package \citep{powell1964efficient,Jones}. While other optimisers may be suitable for this task, as long as they minimize a scalar (negative log likelihood), while varying a vector (relative camera pose), the Powell algorithm achieved acceptable performance with the following tolerance values: $tolx=\num{1e-5}$ and $ftol=\num{1e-8}$. The~objective function that was provided to the optimiser takes the relative camera pose parameters ($\mathbf{t}_c^b = [r_{c,x}^b, r_{c,y}^b, r_{c,z}^b, \theta_c^b e_{c,x}^b, \theta_c^b e_{c,y}^b, \theta_c^b e_{c,z}^b]^T$) and computes the negative log likelihood $-log\Lambda$ (see Equation~(\ref{eq:neg_log_likelihood})) given all pixel locations and navigation system solutions. The optimiser repeatedly calls this function, updating $\mathbf{t}_c^b$ in order to find a relative camera pose $\mathbf{t}_c^{b}\textbf{*}$ that minimises $-log\Lambda$.

As described in Section \ref{sec:uncertainty_estimation} we use \ac{MCMC} to estimate uncertainties in the form of a covariance matrix. \ac{MCMC} was performed with the emcee python package \citep{foreman2013emcee}, which was given a function that computes the log likelihood (Equation (\ref{eq:log_likelihood})). The algorithm was initialised with the previously optimised relative camera pose $\mathbf{t}_c^{b}\textbf{*}$, and run with 250 walkers and 100 iterations, yielding 25,000~samples. A burn in run was also performed with 100 iterations to allow the function to explore the local region prior to performing the actual sampling run. Each sample represents one hypothetical parameter vector $\mathbf{t}_c^b$. The distribution of the samples generated by the \ac{MCMC} algorithm correspond to $log\Lambda$, so the uncertainty of the relative camera pose estimate, $\mathbf{Q}_{\mathbf{t}_c^{b}\textbf{*}}$, can be estimated by computing Equation (\ref{eq:camera_pose_covariance}).

For all calculations of $log\Lambda$, 6$\times$6 covariances for the platform pose were provided by the navigation system at each time stamp. It was assumed that a $u$ pixel point location could be estimated to within one standard deviation of 0.5 pixels (i.e., $\sigma_u=0.5 \,pixels$). If a point is visible it must also be within the \ac{IFOV} of the sensor (see \ref{tab:configuration}), which is approximately 2 pixels for both platforms. We~assumed this to span two standard deviations (95\%), and so one standard deviation is 0.5 pixels ($\sigma_v=0.5 \,pixels$). Principal point and focal length were assumed to have a standard deviation of 2~pixels and 0.1 mm respectively. As previously mentioned, the uncertainty of the relative camera pose, $\mathbf{Q}_{\mathbf{t}_c^{b}}$, was temporarily set to zero (see Sections \ref{sec:cal_pattern_point_estimation} and \ref{sec:reproj_error_and_likelihood}).

\subsection{Outlier Removal} \label{sec:method_outlier_removal}
Unusually high reprojection errors were removed by an iterative process of outlier rejection. First~optimisation was performed on all observations shown in Figure \ref{fig:all_runs} for each platform. Reprojection~errors, $e_{k,i}$ were calculated for each observation $i$ of each pattern point $k$ (see Equation~(\ref{eq:reproj_error})). These~were then averaged per observation:

\begin{equation}
e_i = \frac{\sum_{k}^{M} e_{k,i}}{M}.
\end{equation}

The observation $i$ with the largest mean reprojection error was then removed from the data set and the process was repeated several times (i.e., optimise, calculate reprojection error, remove observation with largest reprojection error). The removal process may be stopped once all mean reprojection errors are below a threshold.

\subsection{Mapping} \label{sec:method_projection}
To demonstrate mapping performance, rays were projected to a plane that was fitted to the estimated pattern point coordinates. Utilising the method in Section \ref{sec:cal_pattern_point_estimation}, the pattern points ($\mathbf{p}^w_k=~[r^w_{k,x},r^w_{k,y},r^w_{k,z}]^T$) were first calculated given the data and relative camera pose . Using the general form of the equation of a plane $ax+by+cz+d=0$, a best fit plane can be found in a linear least squares fashion (setting $c=-1$):

\begin{equation}
\mathbf{A} \mathbf{x} = \mathbf{b},
\hspace{35pt}
\begin{bmatrix} 
r^w_{1,x} & r^w_{1,y} & 1 \\ r^w_{2,x} & r^w_{2,y} & 1 \\ \vdots & \vdots & \vdots \\ r^w_{M,x} & r^w_{M,y} & 1 
\end{bmatrix} 
\begin{bmatrix} a \\ b \\ d \end{bmatrix} = 
\begin{bmatrix} 
r^w_{1,z} \\ r^w_{2,z} \\ \vdots \\ r^w_{M,z}
\end{bmatrix} ,
\end{equation}
\textls[-15]{where the plane parameters $\mathbf{x}$ can be solved for by left multiplying $\mathbf{b}$ with the pseudo-inverse of $\mathbf{A}$, $\mathbf{A}^+$.}

The rays for observation $i$ of pattern point $k$, defined by $\mathbf{p}^w_{c_{i,k}}$ and $\mathbf{p}^w_{s_{i,k}}$ as calculated in Section \ref{sec:1d_cam_model}, can then be projected to the plane by computing their point of intersection:

\begin{equation} \label{eq:plane_projection}
\mathbf{p}^w_{proj_{i,k}}=\frac{([0, 0, -d/c]^T - \mathbf{p}^w_{c_{i,k}}) \cdot [a, b, c]^T} {(\mathbf{p}^w_{s_{i,k}} - \mathbf{p}^w_{c_{i,k}}) \cdot [a, b, c]^T}.
\end{equation} 

Knowing all input covariance matrices, including the covariance of the relative camera position and orientation, as obtained using \ac{MCMC} (see Section \ref{sec:uncertainty_estimation}), the uncertainty of each point projection $\mathbf{p}^w_{proj,i,k}$ may also be calculated. First the partial derivatives of Equation (\ref{eq:plane_projection}) with respect to all inputs were computed to yield the Jacobian $\mathbf{J}_{p_{proj,i,k}}$ of the $x$, $y$ and $z$ position of each point. The uncertainty of each projected point can then be calculated:

\begin{equation}
\mathbf{\Sigma}_{\mathbf{p}^w_{proj_{i,k}}} = \mathbf{J}_{p_{proj_{i,k}}} \mathbf{Q}_{k,i} \mathbf{J}_{p_{proj_{i,k}}}^T,
\end{equation}
where

\begin{equation} \label{eq:proj_cov_matrix}
\mathbf{Q}_i = 
\left[ 
\begin{array}{cccc}
\mathbf{Q}_{uv,k,i} & \mathbf{0} & \mathbf{0} & \mathbf{0} \\
\mathbf{0} & \mathbf{Q}_{\mathbf{t}_{b,k,i}^w} & \mathbf{0} & \mathbf{0} \\
\mathbf{0} & \mathbf{0} & \mathbf{Q}_{int} & \mathbf{0} \\
\mathbf{0} &\mathbf{0} & \mathbf{0} & \mathbf{Q}_{\mathbf{t}_{c}^b} 
\end{array}
\right].
\end{equation}

\subsection{Comparing Poses} \label{sec:method_comparing_poses}
To assess how close a solution is to the optimal, we require a distance metric between two different 6 \ac{DOF} pose transforms. As described in Section \ref{sec:rotation_matrix}, each pose vector is composed of three translation and three orientation parameters. Given two unique poses $\mathbf{t_1} = [r_{1,x}, r_{1,y}, r_{1,z}, \theta_1 e_{1,x}, \theta_1 e_{1,y}, \theta_1 e_{1,z}]^T$ and $\mathbf{t_2} = [r_{2,x}, r_{2,y}, r_{2,z}, \theta_2 e_{2,x}, \theta_2 e_{2,y}, \theta_2 e_{2,z}]^T$, we can compare the translation parts easily by computing their Euclidean distance:

\begin{equation} \label{eq:euclidean_distance}
d_{1,2} = \sqrt{ (r_{2,x}-r_{1,x})^2 + (r_{2,y}-r_{1,y})^2 + (r_{2,z}-r_{1,z})^2  }.
\end{equation}

However, measuring the distance or difference between two rotations is more complicated, and the readers are referred to \citet{Huynh2009} for an in-depth analysis of the topic. \citet{Huynh2009} presents  and recommends a number of metrics for comparing rotations. Of these, the geodesic on the unit sphere was chosen, because it represents the magnitude of the rotation angle required to align the two rotations, which was deemed to be an intuitive measure. It can be computed as follows:

\begin{equation} \label{eq:product_of_quaternions}
\Phi_{1,2} = 2\arccos(|\mathbf{q}_1 \cdot \mathbf{q}_2|),
\end{equation}
where $\mathbf{q}_1$ and $\mathbf{q}_2$ are unit quaternion equivalents of $[\theta_1 e_{1,x}, \theta_1 e_{1,y}, \theta_1 e_{1,z}]^T$ and $[\theta_2 e_{2,x}, \theta_2 e_{2,y}, \theta_2 e_{2,z}]^T$ respectively, computed as:

\begin{equation} \label{eq:axis_angle_to_quat}
\begin{split}
\mathbf{q}_i = & a_i + b_i \mathbf{i} + c_i \mathbf{j} + d_i \mathbf{k} \\ 
= & \cos \left(\frac{\theta_i}{2} \right) + e_{i,x}\sin \left(\frac{\theta_i}{2} \right) \mathbf{i} + \\
& e_{i,y}\sin \left(\frac{\theta_i}{2} \right) \mathbf{j} + e_{i,z}\sin \left(\frac{\theta_i}{2} \right) \mathbf{k}.
\end{split}
\end{equation}

$\Phi_{1,2}$ can also be interpreted to be equal to the absolute value of the angular magnitude $\theta$ (in the range $[-\pi,\pi]$) of the axis-angle rotation required to align the two orientations. For this reason, the metric will be simply referred to as the axis-angle difference. Combined, $d_{1,2}$ and $\Phi_{1,2}$ form a 2D pose distance that is convenient for visualisation.

\subsection{Basin of Attraction} \label{sec:method_basin_of_attraction}
Because the method in this paper requires an approximate initial camera pose, it is important to numerically quantify how precise this initial camera pose must be to yield an accurate optimised estimate. To measure how far an initial hand measured camera pose can be from the optimum, while still resulting in correct global convergence, we test a set of starting conditions that are altered by different amounts, and measure how close the optimal result is from the known global optimum. Due~to the high dimensionality of the search space, a random sub-sampling of initial poses is performed. Deviation of initial values from the known optimum is quantified in the two dimensional pose distance space defined in Section \ref{sec:method_comparing_poses}. The 2D space was sampled uniformly in a grid, and for each location a 6D initial parameter vector was randomly generated at the corresponding Euclidean and axis-angle distance from the known optimal value. 

First an $x$, $y$ and $z$ translation vector was generated at random, uniformly distributed over an equal negative to positive range for all three parameters. The vector was normalised to unity and then multiplied by the Euclidean distance value of the corresponding grid position. The resulting vector was added to the optimised translation parameters, yielding the initial coordinates. Similarly, three axis-angle orientation values were randomly generated in the same way, normalised to unity and multiplied by the corresponding axis-angle difference value at the  given grid location, yielding an axis-angle rotation. The optimised orientation parameters were then rotated by this difference rotation, producing the initial orientation values. 

This yields a set of sparse random 6 \ac{DOF} samples that are uniformly spaced in terms of pose distance from the known optimal camera pose, allowing the basin of attraction to be mapped. Optimisation was performed for each randomly generated initial camera pose on the grid. For~each result, the Mahalanobis distance to the reference optimum was calculated, given the covariance matrix resulting from the \ac{MCMC} uncertainty estimation on the optimised parameters. 

\section{Results} \label{sec:results}

This section presents the results of line scan camera pose estimation for two different platforms and configurations, including outlier rejection, resulting camera pose and uncertainty, in-depth analysis of the uncertainty, the impact of platform pose diversity on the accuracy, and finally the combined mapping uncertainty.

\subsection{Outlier Rejection} 

\begin{figure}[H]
	\centering
	\begin{subfigure}{0.42\textwidth}
		\centering
		\includegraphics[width=1.\textwidth]{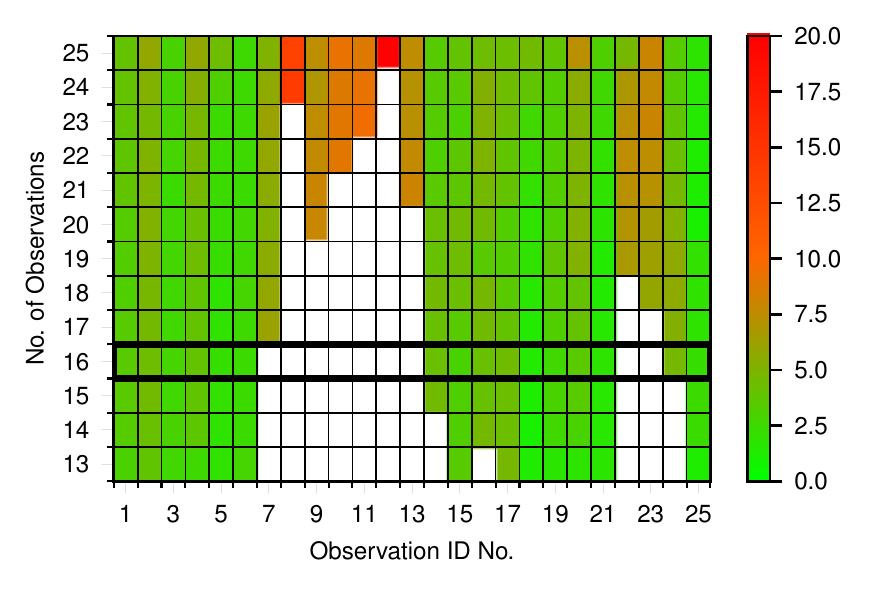}	
		\caption{Ladybird}
		\label{fig:outlier_removal_ladybird}
	\end{subfigure}
	\begin{subfigure}{0.42\textwidth}
		\centering
		\includegraphics[width=1.\textwidth]{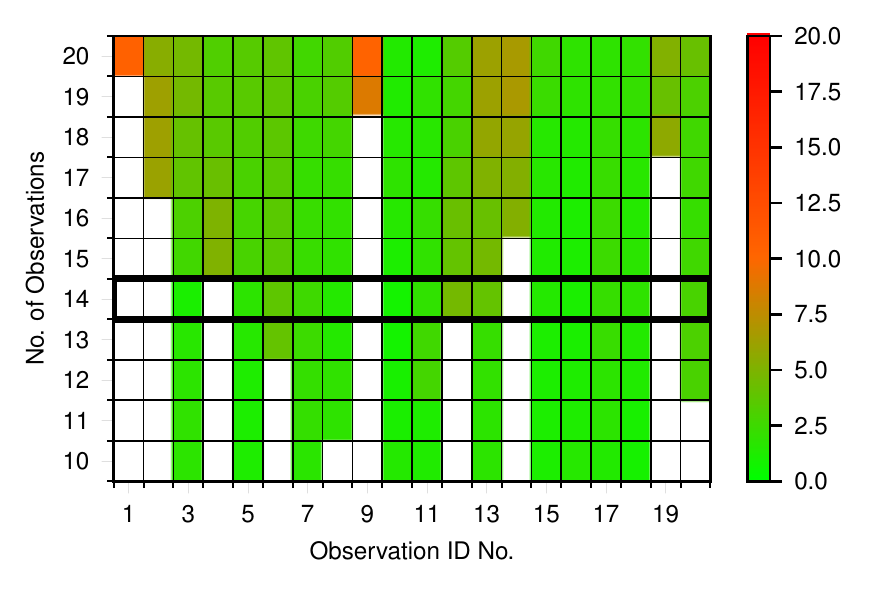}
		\caption{Shrimp}
		\label{fig:outlier_removal_shrimp}
	\end{subfigure}
	\vspace{-12pt}
	\caption{Outlier removal and average reprojection errors for (\textbf{a}) Ladybird and (\textbf{b}) Shrimp. Each row represents an iteration as described in Section \ref{sec:method_outlier_removal}, labelled with the number of remaining observations, and each column represents an observation of the calibration pattern as per Figure \ref{fig:all_runs}. White cells indicate that an observation has been removed for that iteration. Otherwise, the cell colour indicates the mean reprojection error in pixels for that particular observation of the calibration pattern at a given outlier removal iteration. For example, in (\textbf{a}), observation No. 12 exhibited the greatest mean reprojection error with 25 observations, and was therefore the first observation to be removed. The~highlighted rectangle points to the row with the greatest number of remaining observations with a mean reprojection error of less than 5 pixels.}
	\label{fig:outlier_removal}
\end{figure}

The iterative results of outlier removal based on reprojection errors are shown in Figure \ref{fig:outlier_removal}. Each~row represents an outlier removal iteration, labelled by the number of remaining observations, where the top row includes all observations. Each column represents one of the observations from Figure~\ref{fig:all_runs}, where each observation is one view of all 15 points on the calibration pattern. The colour of each cell indicates the mean reprojection error of the 15 points within the single observation of the calibration pattern, for a particular outlier removal iteration. In each figure, the black rectangle highlights the row with the greatest number of observations that all have mean errors less than a 5~pixel threshold. Ladybird exhibited a greater number of outliers and higher worst-case reprojection errors than Shrimp, with 9~compared to 6 outliers respectively. This resulted in 16 inliers for Ladybird and 14 for Shrimp.

\subsection{Pose and Uncertainty Results} 
The relative camera pose transforms and associated uncertainties (one standard deviation) for both platforms are shown in Table \ref{tab:results} before and after outlier removal. The hand measured estimate is also shown for reference, where the tolerances reflect the difficulty of measurement. The results for the Shrimp platform exhibit greater uncertainty compared to the Ladybird platform.

\begin{table}[H]
		\tablesize{\footnotesize}
		\centering
		\caption{\boldmath{$t_c^b$} estimates and uncertainties (axis-angle rotations).}
		\tabcolsep=0.11cm
		\begin{tabulary}{1\textwidth}{lccccccccccc}
		\toprule
			 & \boldmath{$r_{c,x}^b$} \textbf{(m)} & \boldmath{$r_{c,y}^b$} \textbf{(m)} & \boldmath{$r_{c,z}^b$} \textbf{(m)} & \boldmath{$\theta_c^b e_{c,x}^b$} \textbf{(rad)} & \boldmath{$\theta_c^b e_{c,y}^b$} \textbf{(rad)} & \boldmath{$\theta_c^b e_{c,z}^b$} \textbf{(rad)} \\
\hline \noalign{\vskip 1mm} 
\textbf{Ladybird} &  &  &  &  &  &  \\
\midrule
All observations & $0.147 \pm 0.025$ & $-0.128 \pm 0.045$ & $-0.630 \pm 0.051$ & $-0.849 \pm 0.013$ & $0.768 \pm 0.015$ & $-1.420 \pm 0.008$ \\
16 observations & $0.189 \pm 0.032$ & $-0.142 \pm 0.054$ & $-0.794 \pm 0.057$ & $-0.822 \pm 0.016$ & $0.738 \pm 0.018$ & $-1.429 \pm 0.009$ \\
Hand measured & $0.200 \pm 0.100$ & $0.000 \pm 0.100$ & $-0.800 \pm 0.100$ & $-0.762 \pm 0.039$ & $0.762 \pm 0.039$ & $-1.433 \pm 0.037$ \\
 \midrule
\textbf{Shrimp} &  &  &  &  &  &  \\
\midrule
All observations & $0.044 \pm 0.031$ & $-0.133 \pm 0.096$ & $-0.660 \pm 0.158$ & $1.409 \pm 0.018$ & $1.400 \pm 0.027$ & $-1.078 \pm 0.019$ \\
14 observations & $-0.010 \pm 0.069$ & $-0.080 \pm 0.141$ & $-0.579 \pm 0.178$ & $1.380 \pm 0.030$ & $1.427 \pm 0.042$ & $-1.093 \pm 0.026$ \\
Hand measured & $0.000 \pm 0.100$ & $-0.200 \pm 0.100$ & $-0.500 \pm 0.100$ & $1.399 \pm 0.026$ & $1.399 \pm 0.088$ & $-1.074 \pm 0.071$ \\

			\bottomrule
		\end{tabulary}\\
		\begin{tabular}{@{}c@{}} 
\multicolumn{1}{p{\textwidth -.88in}}{\footnotesize Note: Hand measured orientation uncertainties are 2\degree  \,for each parameter in Euler representation, converted to axis-angle representation by propagating the covariance matrix using the Jacobians of the conversion function.}
\end{tabular}

\label{tab:results}
\end{table}

\begin{figure}[H]
	\centering
	\begin{subfigure}{0.49\textwidth}
		\centering
		\includegraphics[width=1.\textwidth]{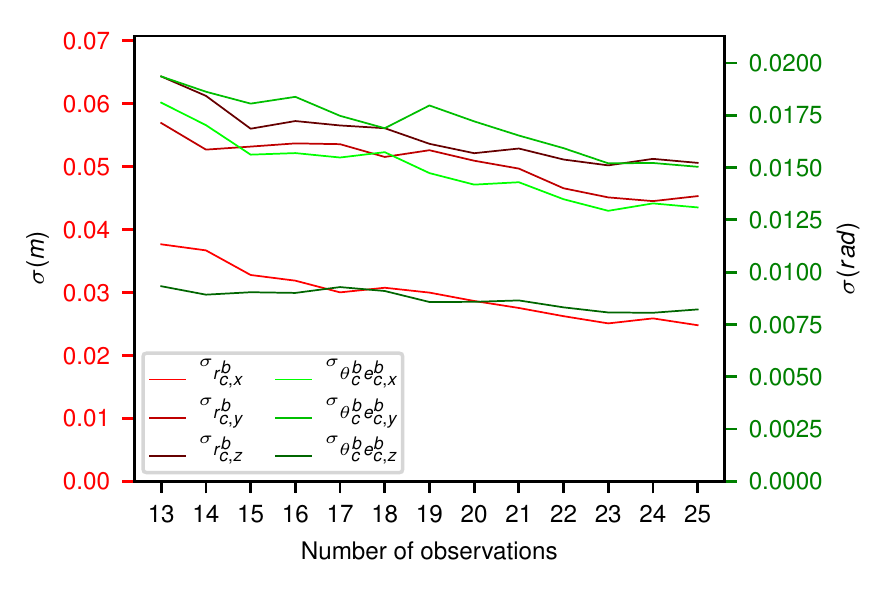}	
		\caption{Ladybird}
		\label{fig:uncertainty_increase_ladybird}
	\end{subfigure}
	\begin{subfigure}{0.49\textwidth}
		\centering
		\includegraphics[width=1.\textwidth]{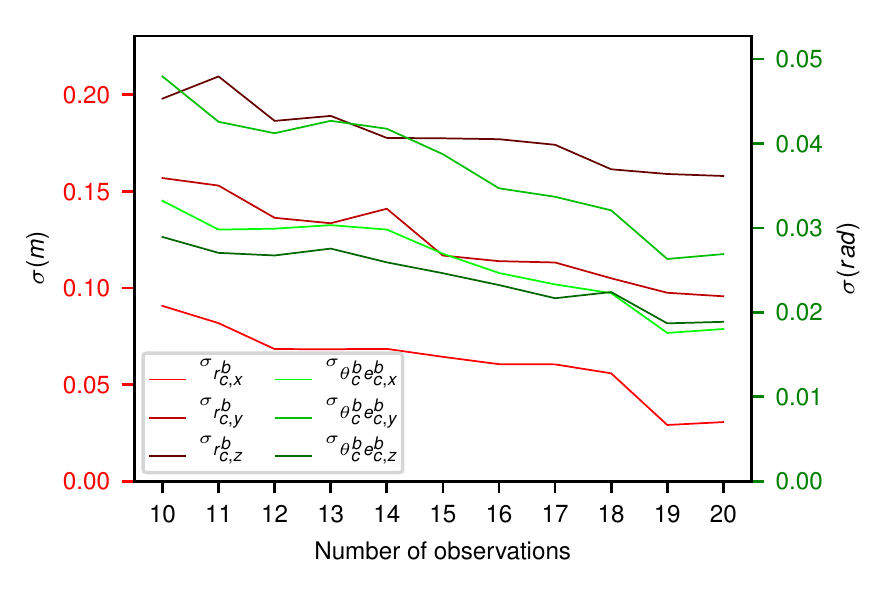}
		\caption{Shrimp}
		\label{fig:uncertainty_increase_shrimp}
	\end{subfigure}
	\vspace{-12pt}
	\caption{The number of observations vs. standard deviation for (\textbf{a}) Ladybird and (\textbf{b}) Shrimp respectively. The x axis mirrors the outlier removal stages shown in Figure \ref{fig:outlier_removal}.}
	\label{fig:uncertainty_increase}
\end{figure}

In Figure \ref{fig:uncertainty_increase} each outlier removal stage is plotted against each parameter's standard deviation. As~would be expected, increasing the number of observations decreases the uncertainty of the estimate.

The number of observations affects the number of computations and therefore has a significant effect on calibration and \ac{MCMC} run time. For Ladybird, optimisation and \ac{MCMC} took approx. 15~min and 7 h respectively for all 25 observations. For 16 observations, this was reduced to just over 7~min and 4 h. For Shrimp, the respective optimisation and \ac{MCMC} times were reduced from approx. 5.5~min and 5 h for all 20 observations to just over 2 min and 3 h with 14 observations. 

\subsection{In-Depth Uncertainty Analysis} \label{sec:results_uncertainty_analysis}
\begin{figure}[H]
	\centering
	\begin{subfigure}{0.49\textwidth}
		\includegraphics[width=1.\textwidth] {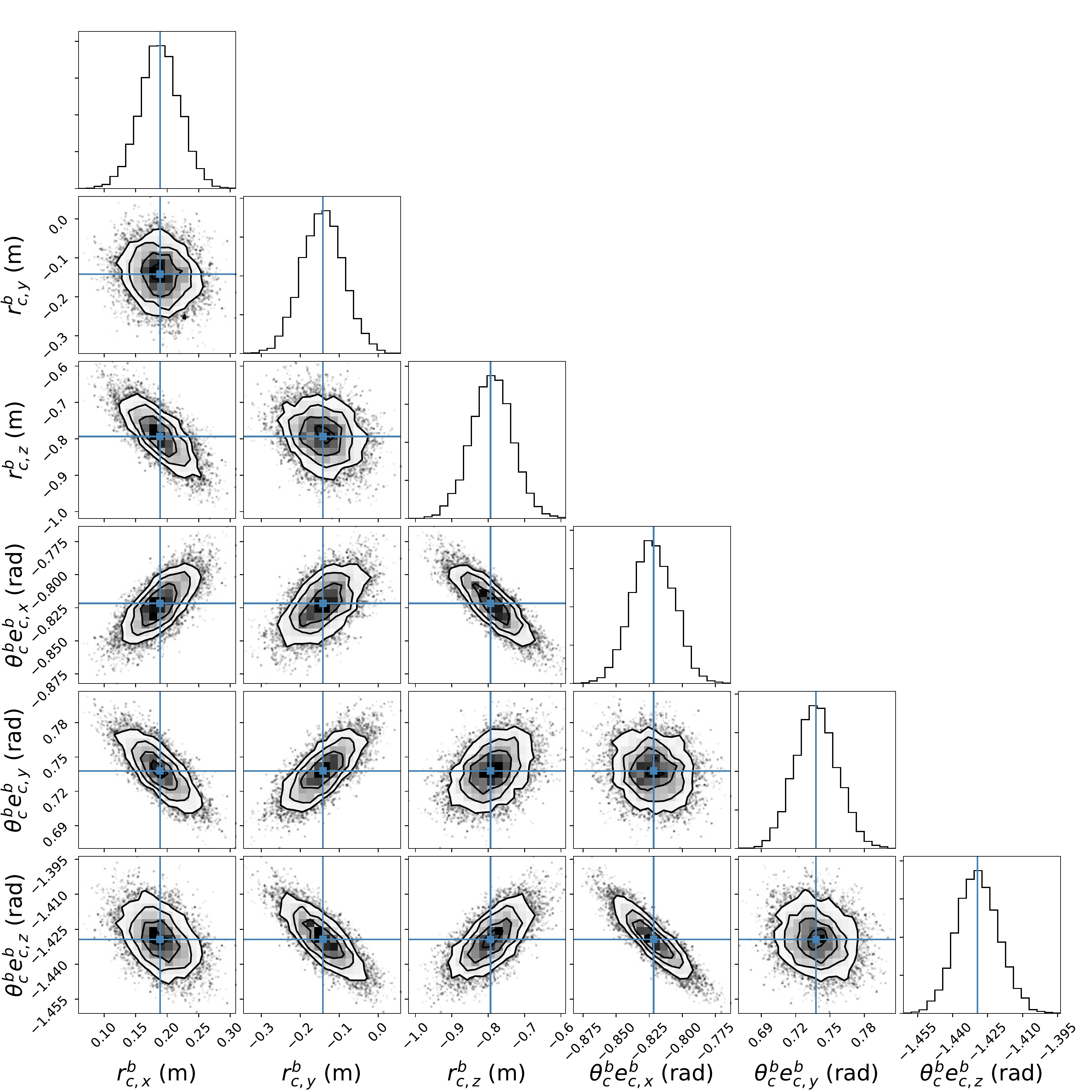}	
		\caption{Ladybird}
		\label{fig:corner_ladybird}
	\end{subfigure}
	\begin{subfigure}{0.49\textwidth}
		\includegraphics[width=1.\textwidth]{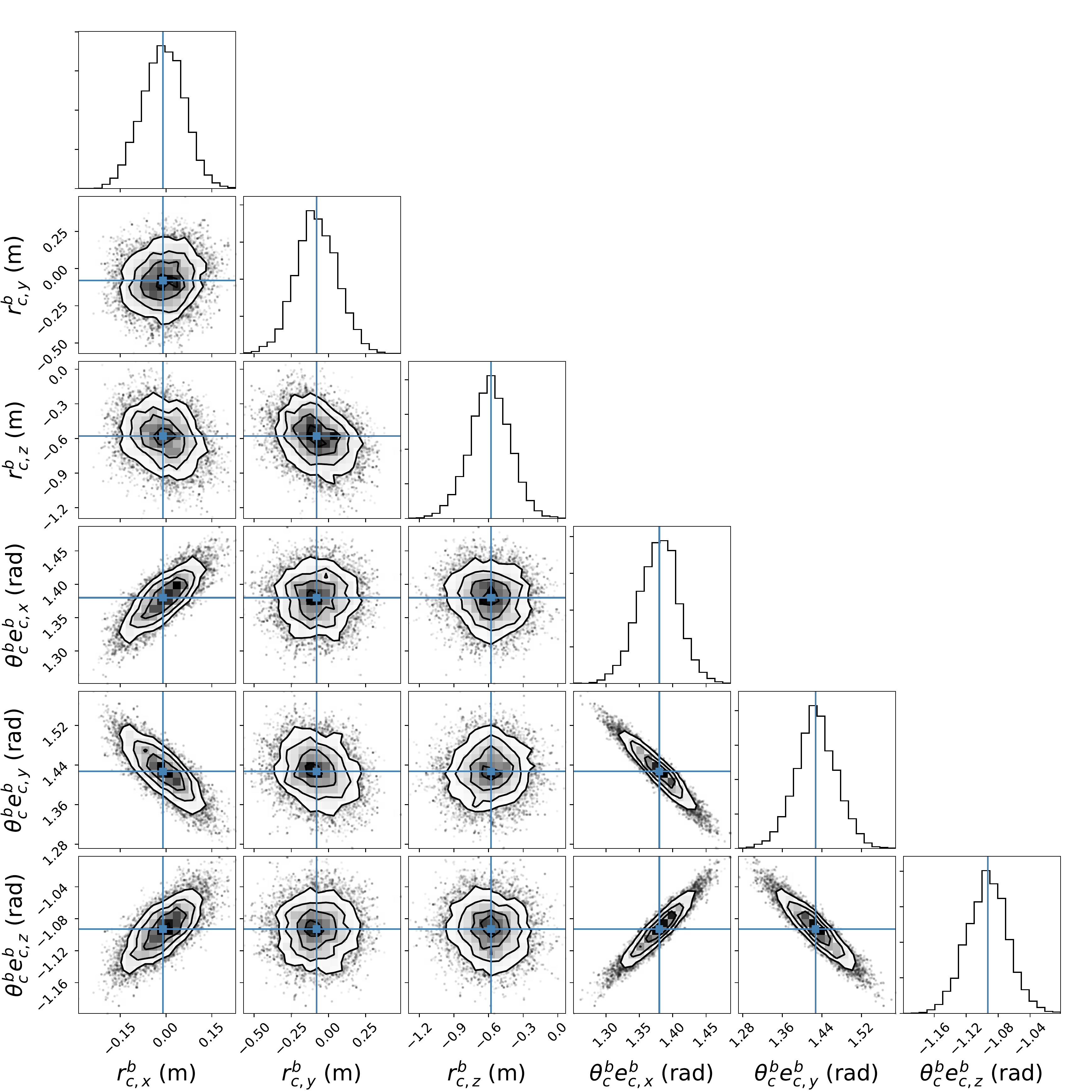}
		\caption{Shrimp}
		\label{fig:corner_shrimp}
	\end{subfigure}
	\vspace{-12pt}
	\caption{\textls[-5]{Corner plots for (\textbf{a}) Ladybird and (\textbf{b}) Shrimp platforms for 16 and 14 observations respectively. Each sub-plot below the diagonal provides a 2D histogram of \ac{MCMC} sample values for a pair of relative camera pose parameters. The sub-plots on the diagonal show 1D histograms for each parameter.}}
	\label{fig:corner_plot}
\end{figure}

\textls[-25]{Examining uncertainty in more detail, \ac{MCMC} samples are shown on a corner plot in Figure \ref{fig:corner_plot} \citep{corner}.} Each sub-plot below the diagonal provides a 2D histogram, showing the \ac{MCMC} sample density between two parameters (i.e., the marginal likelihood distribution for a parameter pair), and on the diagonal a 1D histogram, giving the sample density for the marginal likelihood distribution for each single parameter.

\begin{figure}[H]
	\centering
	\begin{subfigure}{0.3\textwidth}
		\includegraphics[width=1.\textwidth,trim={0 15mm 0 0},clip] {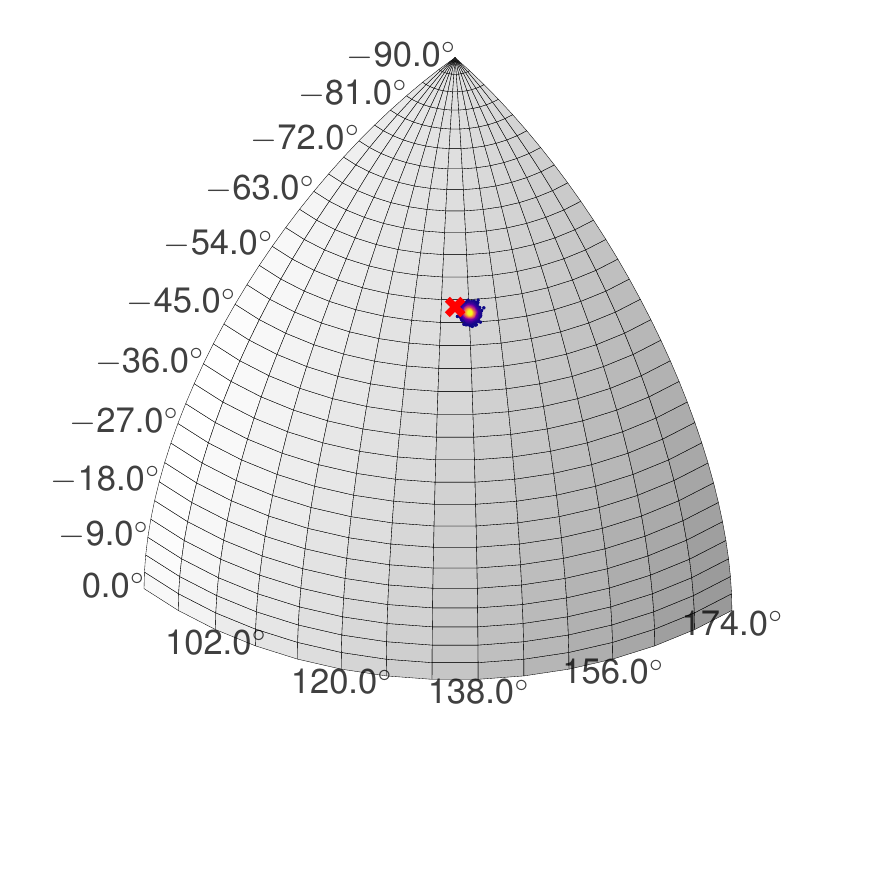}	
		\caption{Ladybird sphere}
		\label{fig:sphere_ladybird}
	\end{subfigure}
	\begin{subfigure}{0.28\textwidth}
		\includegraphics[width=1.\textwidth]{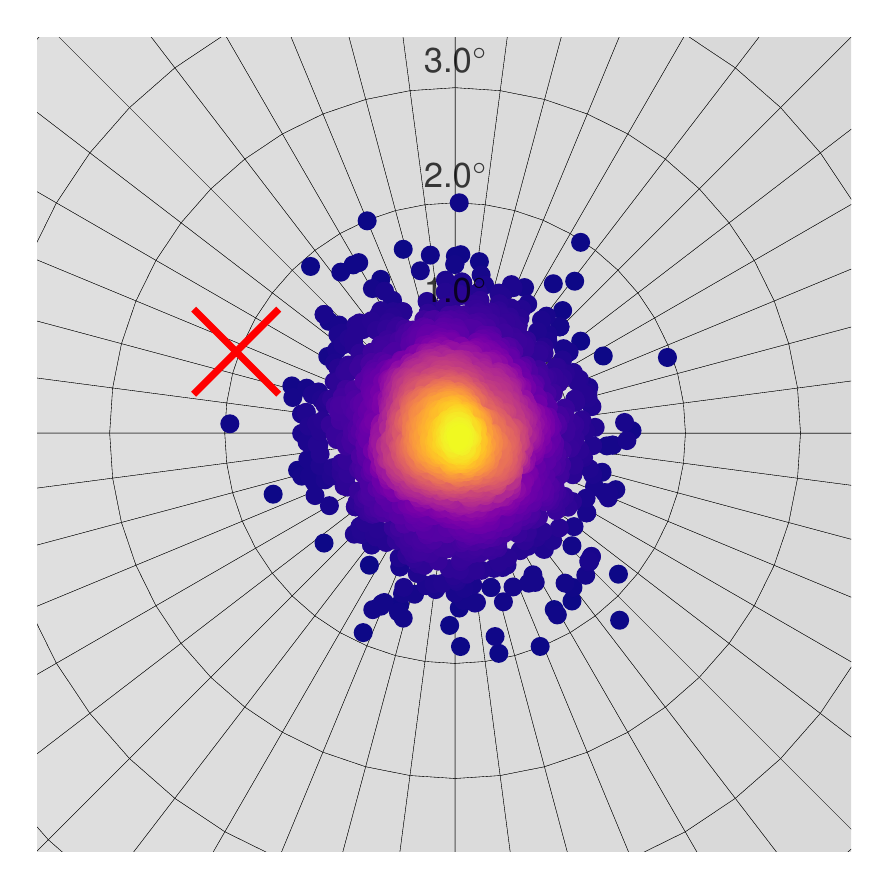}
		\caption{Ladybird sphere close-up}
		\label{fig:sphere_ladybird_close}
	\end{subfigure}
	\begin{subfigure}{0.4\textwidth}
		\includegraphics[width=1.\textwidth]{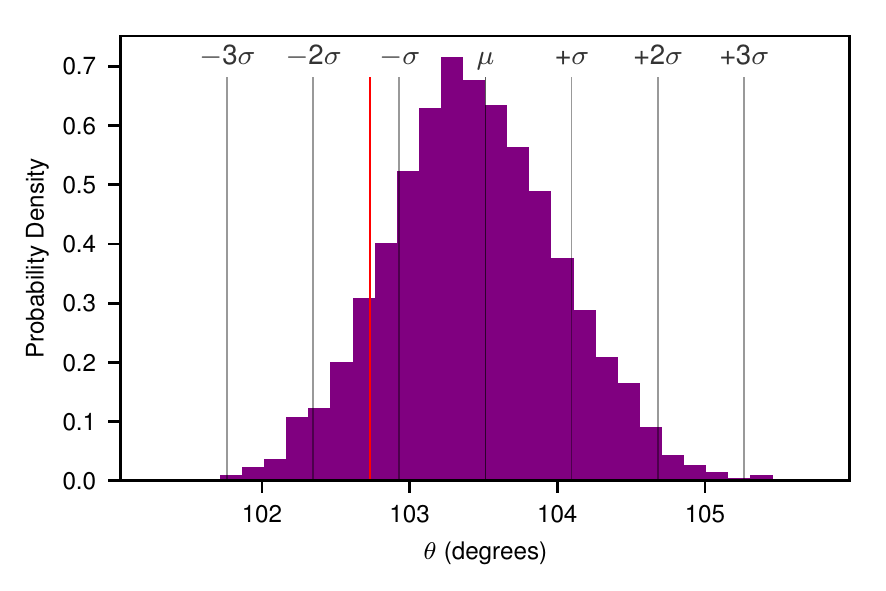} 
		\caption{Ladybird axis-angle magnitude}
		\label{fig:aa_hist_ladybird}
	\end{subfigure}
	\begin{subfigure}{0.3\textwidth}
		\includegraphics[width=1.\textwidth,trim={0 15mm 0 0},clip] {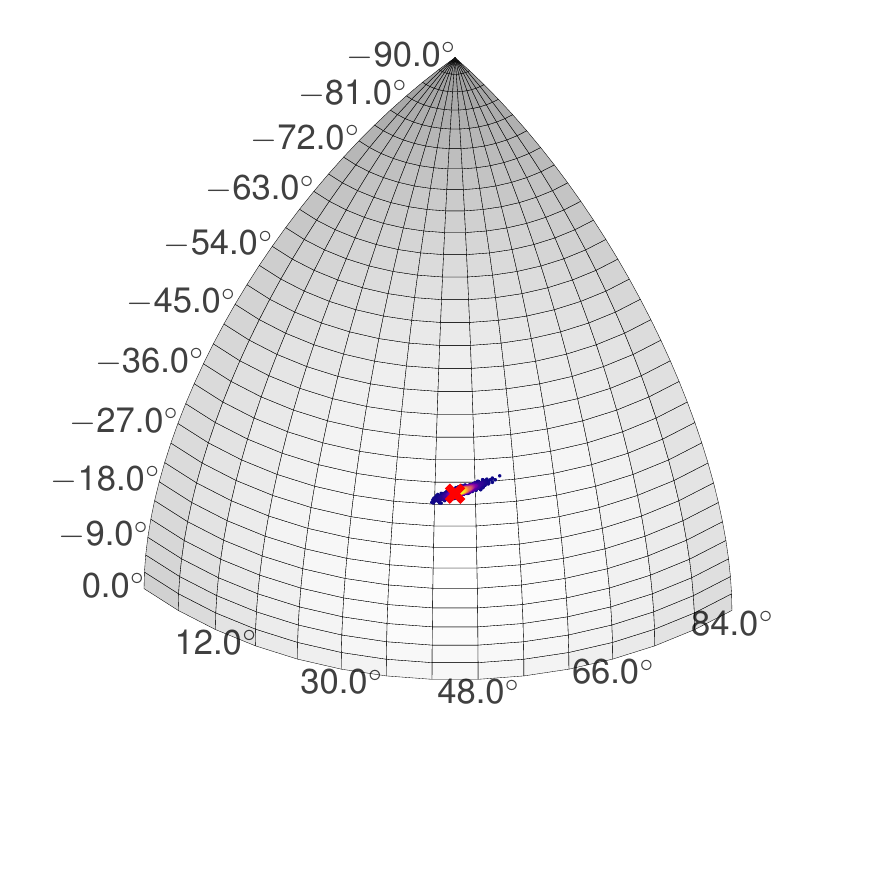}	
		\caption{Shrimp sphere}
		\label{fig:sphere_shrimp}
	\end{subfigure}
	\begin{subfigure}{0.28\textwidth}
		\includegraphics[width=1.\textwidth]{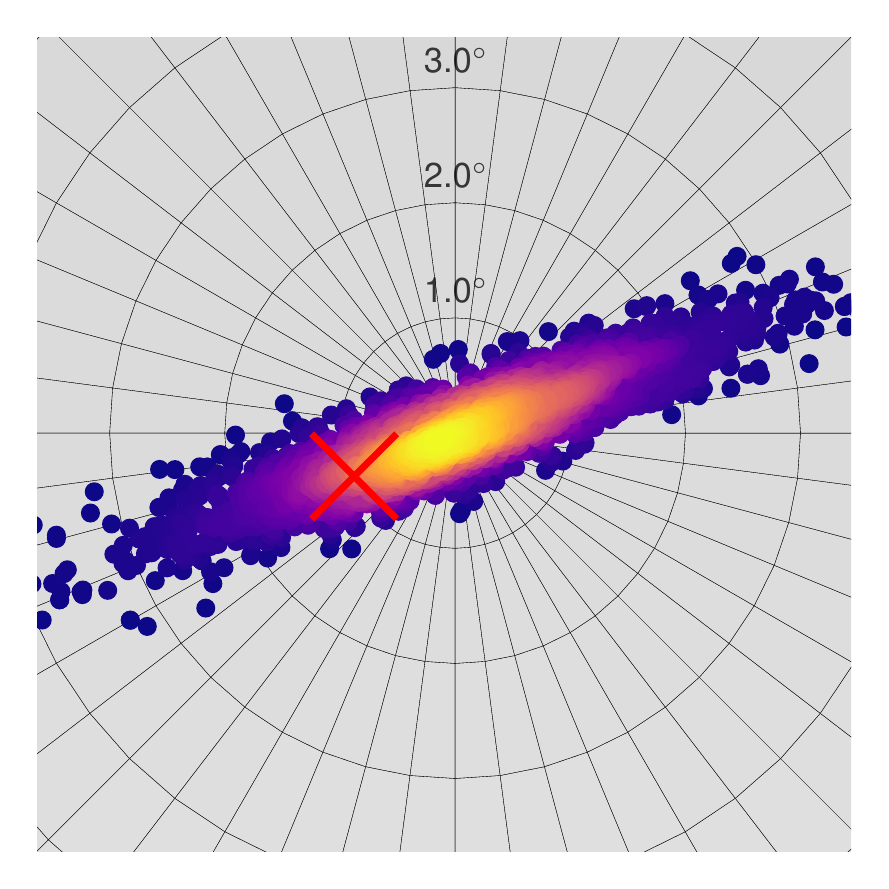}
		\caption{Shrimp sphere close-up}
		\label{fig:sphere_shrimp_close}
	\end{subfigure}
	\begin{subfigure}{0.4\textwidth}
		\includegraphics[width=1.\textwidth]{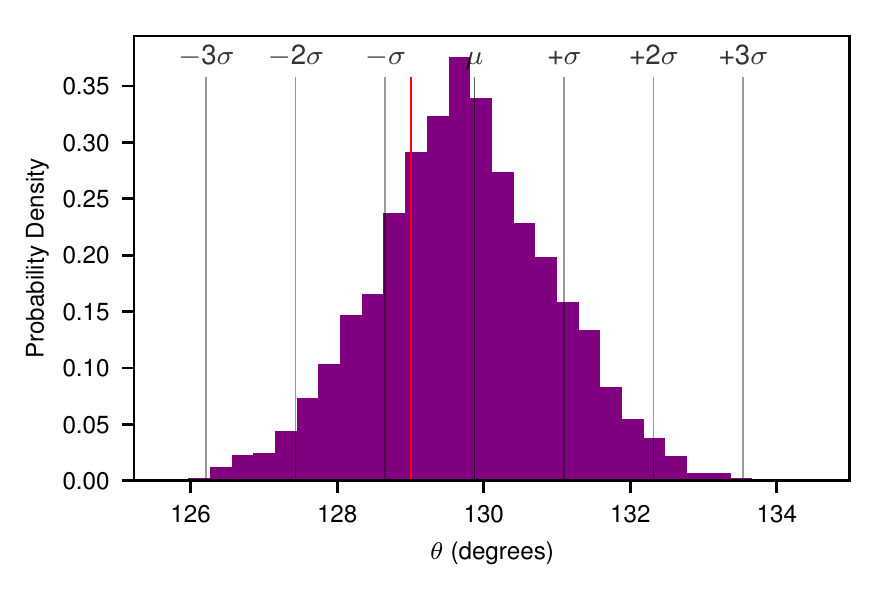} 
		\caption{Shrimp axis-angle magnitude}
		\label{fig:aa_hist_shrimp}
	\end{subfigure}
	\vspace{-12pt}
	\caption{Ladybird and Shrimp platform \ac{MCMC} axis-angle unit vector ($\mathbf{e_s^b}$) samples plotted on a sphere (\textbf{a},\textbf{d}) and close-up, centred to a pole (\textbf{b},\textbf{e}) (using 16 and 14 observations for Ladybird and Shrimp respectively). Each point on the sphere is coloured according to a \ac{KDE} to give an indication of the axis-angle vector marginal likelihood for visualisation purposes only. Points were randomly thinned by ten to facilitate plotting. A histogram of the magnitudes of the axis-angle rotation, $\theta_s^b$, is shown in (\textbf{c},\textbf{f}). The hand measured pose orientation is shown with red crosses in (\textbf{a},\textbf{b},\textbf{d},\textbf{e}), and a vertical red line in (\textbf{c},\textbf{f}).}
	\label{fig:spheres}
\end{figure}

For human interpretation of the uncertainty, in Figure \ref{fig:spheres}a,d the \ac{MCMC} sample values for $[\theta e_{c,x}^b, \theta e_{c,y}^b, \theta e_{c,z}^b]^T$ are plotted on a sphere. For visualisation only, each point is coloured according to a \ac{KDE} performed on all \ac{MCMC} samples to give an indication of the axis-angle vector marginal likelihood. Hand measured pose estimates are shown as red crosses. Likewise, in Figure \ref{fig:spheres}c,f, values for $\theta_c^b$ are presented in a histogram, showing the marginal likelihood of the axis-angle magnitude. While both figures indicate a clustering to within two degrees, the Shrimp platform's distribution exhibits a more elongated elliptical shape, while for Ladybird, it is more uniformly spread in all directions. Also apparent for Ladybird is that the manual measurements are well outside the region of high likelihood, both in terms of the axis-angle unit vector and magnitude. Conversely, the hand measured pose for the Shrimp platform is highly likely given the data, suggesting the initial manual measurement may have been more accurate than for Ladybird.

In Figure \ref{fig:result_on_model}, the distributions of \ac{MCMC} samples is shown superimposed on the corresponding platform model. Translation parameters are presented as a 2D histogram, similar to Figure \ref{fig:corner_plot}, demonstrating the marginal density of the likelihood distribution from each orthogonal viewpoint. To~present the orientation parameter distribution, line segments coincident with the camera's viewing direction, and anchored to the optimized camera centre, are rotated by each \ac{MCMC} rotation sample. Each line is semi-transparent, and so as all samples build up, the density distribution of the camera orientation is visualised. It is evident that there is greater variance in the \ac{MCMC} samples for Shrimp when compared to the Ladybird platform, particularly for the translation parameters, as corroborated by the numerical results in Table \ref{tab:results}.

\subsection{Effect of Angular Diversity}
To investigate how angular diversity of body poses in a dataset affects the certainty of the result, two experimental subsets were compiled from the outlier-rejected dataset with 16 observations for Ladybird. The first includes only ten observations with small roll angles $\phi_{b,x}^w < 1.9\degree$. The second includes five observations with small roll $\phi_{b,x}^w$ and five with roll angles $3.3\degree < \phi_{b,x}^w < 5.8\degree$. Both datasets therefore contain ten total observations, representing low and high angular diversity. Optimisation results for these subsets are shown in Table \ref{tab:angular_diversity}. Despite containing the same number of observations, the dataset with high angular diversity results in significantly lower uncertainty for the optimal camera pose, compared to the low diversity set.

\unskip
\begin{table}[H]
	\centering
	\tablesize{\footnotesize}
	\caption{Angular diversity comparison.}
	\tabcolsep=0.11cm
	\begin{tabulary}{0.95\textwidth}{lcccccc}
	\toprule
		 & \boldmath{${r_{c,x}^b}$ \textbf{(m)}} & \textbf{ \boldmath{$r_{c,y}^b$} (m)} & \textbf{\boldmath{$r_{c,z}^b$} (m)} & \textbf{\boldmath{$\theta_c^b e_{c,x}^b$} (rad)} & \textbf{\boldmath{$\theta_c^b e_{c,y}^b$} (rad)} & \textbf{\boldmath{$\theta_c^b e_{c,z}^b$} (rad)} \\
\hline \noalign{\vskip 1mm} 
High ang. diversity & $0.166 \pm 0.041$ & $-0.155 \pm 0.083$ & $-0.717 \pm 0.074$ & $-0.831 \pm 0.023$ & $0.744 \pm 0.023$ & $-1.420 \pm 0.015$ \\
Low ang. diversity & $0.147 \pm 0.118$ & $0.058 \pm 0.107$ & $-0.269 \pm 0.281$ & $-0.797 \pm 0.051$ & $0.796 \pm 0.059$ & $-1.452 \pm 0.026$ \\

		\bottomrule
	\end{tabulary}
	\label{tab:angular_diversity}
\end{table}

\subsection{Repeatability}
To assess the repeatability of the approach, another two experimental subsets were compiled from the outlier-rejected dataset with 16 observations for Ladybird. They each contained five observations with small roll $\phi_{b,x}^w$ and three with roll angles $3.3\degree < \phi_{b,x}^w < 5.8\degree$. The two subsets do not share any observations, making them independent. Optimisation results for these subsets are shown in Table~\ref{tab:repeatability}. The two results are consistent with each other, as the two distributions overlap to a significant degree. Due to the small dataset size, uncertainty values are higher than the result in Table \ref{tab:results} with all 16~observations, which is expected as demonstrated in Section \ref{sec:results_uncertainty_analysis}.

\begin{figure}[H]
	\centering
	\begin{subfigure}{0.49\textwidth}
		\centering
		\includegraphics[width=1.\textwidth]{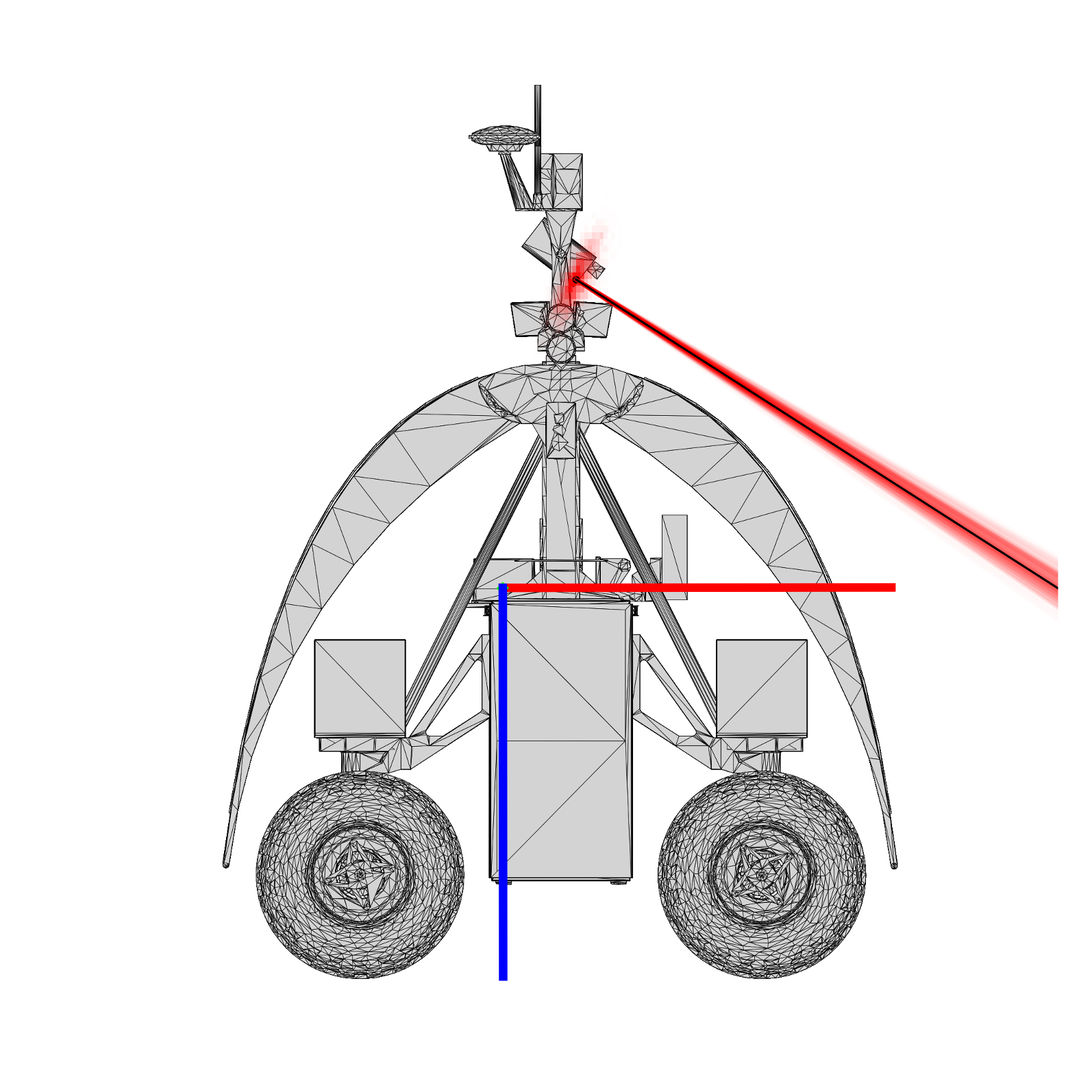}	
		\caption{Ladybird - Side}
		\label{fig:result_on_model_lb_side}
	\end{subfigure}
	\begin{subfigure}{0.49\textwidth}
		\centering
		\includegraphics[width=1.\textwidth]{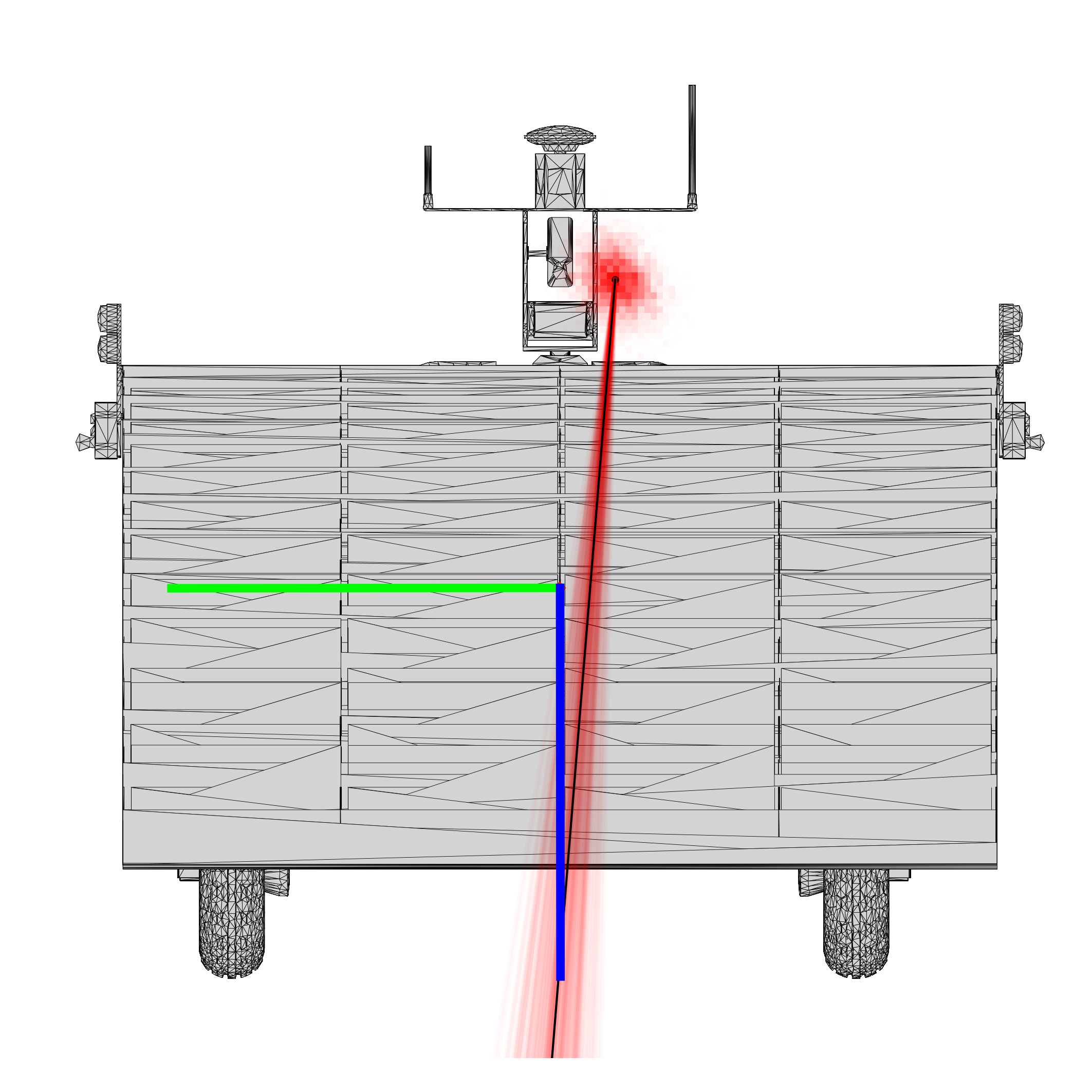}
		\caption{Ladybird - Front}
		\label{fig:result_on_model_lb_front}
	\end{subfigure}
	\begin{subfigure}{0.49\textwidth}
		\centering
		\includegraphics[width=1.\textwidth]{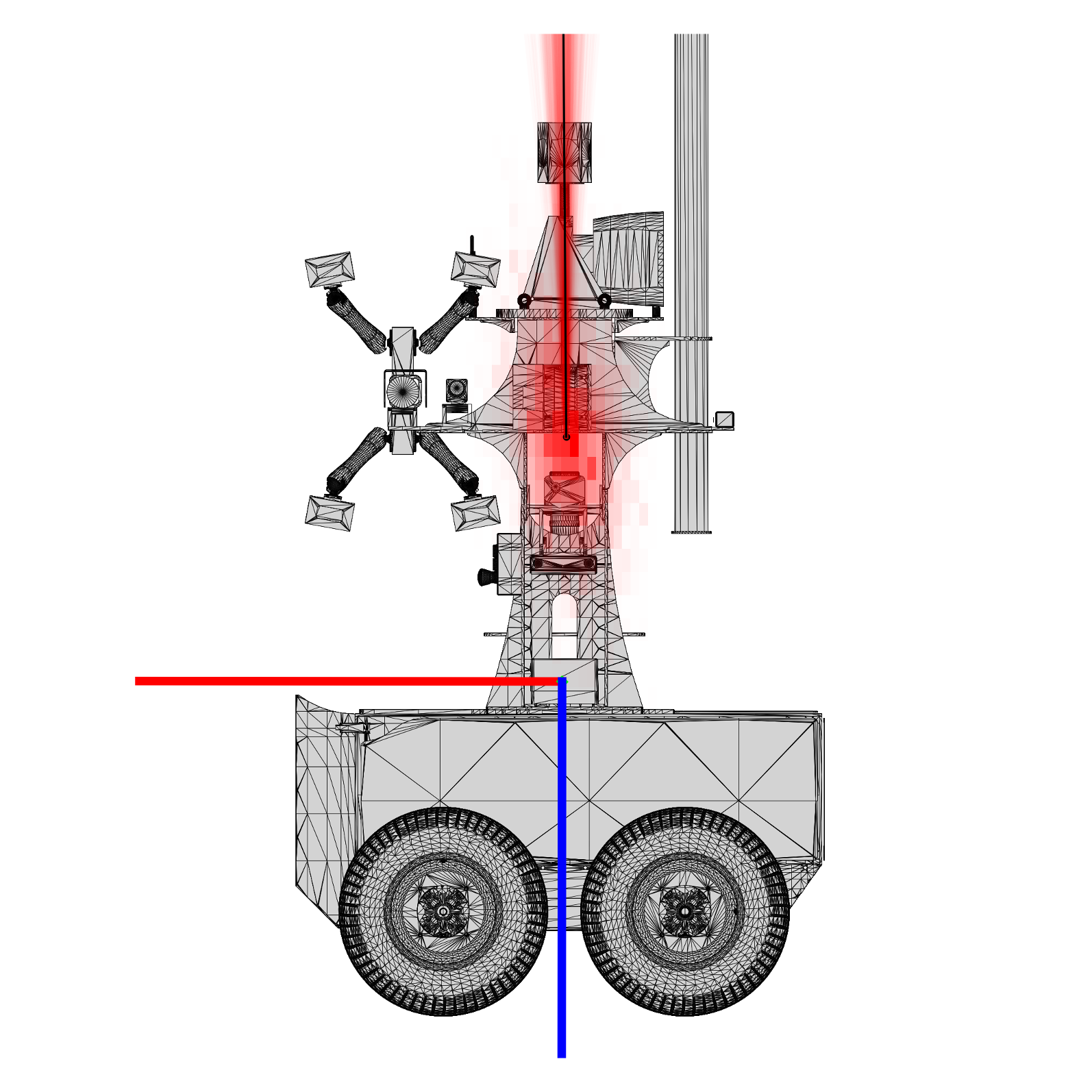}	
		\caption{Shrimp - Side}
		\label{fig:result_on_model_shrimp_side}
	\end{subfigure}
	\begin{subfigure}{0.49\textwidth}
		\centering
		\includegraphics[width=1.\textwidth]{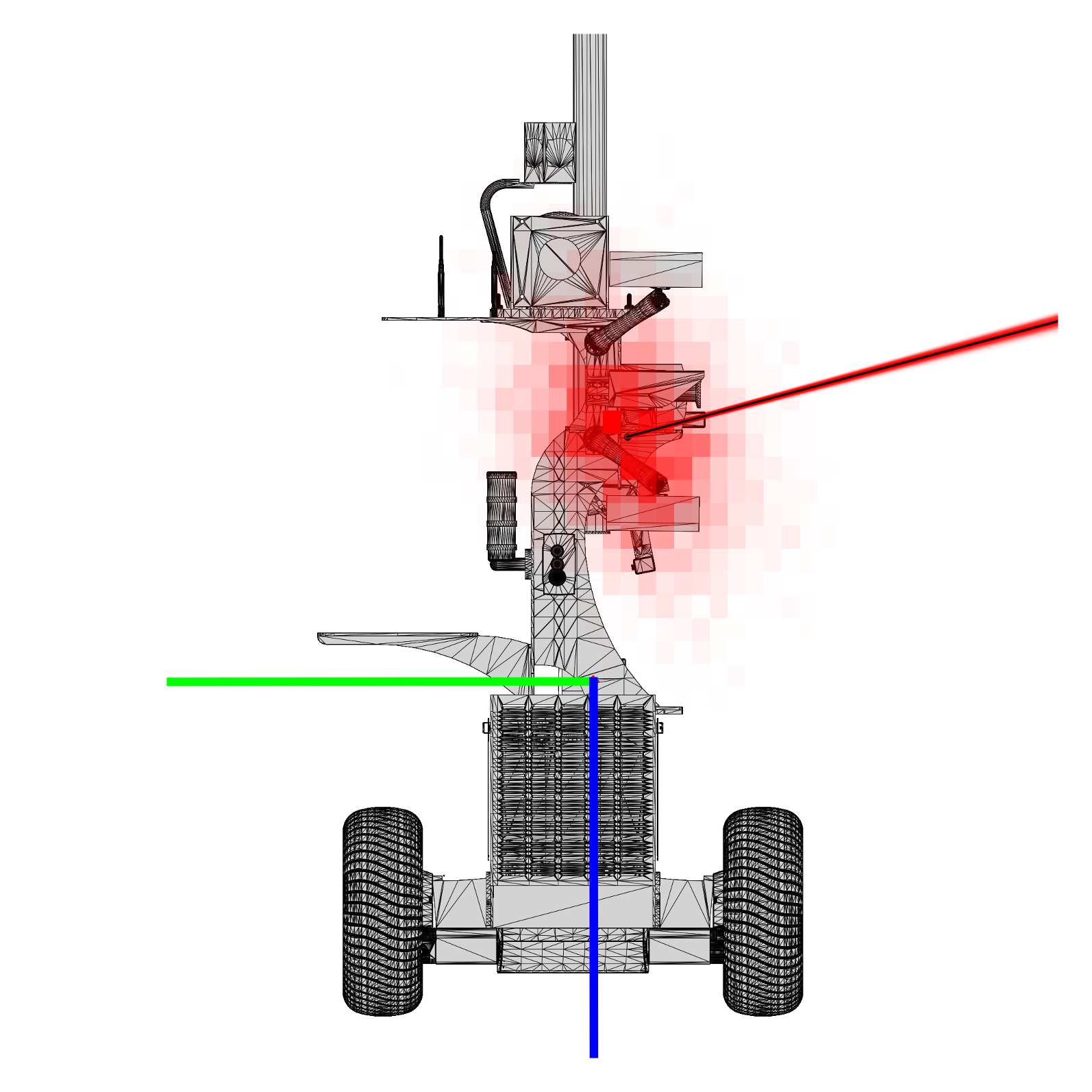}
		\caption{Shrimp - Front}
		\label{fig:result_on_model_shrimp_front}
	\end{subfigure}
	\vspace{-12pt}
	\caption{Optimised camera pose and \ac{MCMC} sample poses for Ladybird (\textbf{a},\textbf{b}) and Shrimp (\textbf{c},\textbf{d}). The~optimised pose is shown as an $xyz$ axis with black lines. The \ac{MCMC} samples are shown as red, green and blue $xyz$ axes, where greater colour intensity corresponds to greater sample density, approximating the marginal likelihood of the poses in the viewing direction. The body (i.e., navigation system) pose is also shown as a red, green and blue $xyz$ axis.}
	\label{fig:result_on_model}
\end{figure}
\unskip
\begin{table}[H]
	\centering
	\tablesize{\footnotesize}
	\caption{Repeatability.}
	\tabcolsep=0.11cm
	\begin{tabulary}{0.95\textwidth}{lcccccc}
	\toprule
		 & \textbf{\boldmath{$r_{c,x}^b$} (m)} & \textbf{\boldmath{$r_{c,y}^b$} (m)} & \textbf{\boldmath{$r_{c,z}^b$} (m)} & \textbf{\boldmath{$\theta_c^b e_{c,x}^b$} (rad) }& \textbf{\boldmath{$\theta_c^b e_{c,y}^b$} (rad)} & \textbf{\boldmath{$\theta_c^b e_{c,z}^b$} (rad)} \\
\hline \noalign{\vskip 1mm} 
Subset 1 & $0.192 \pm 0.046$ & $-0.170 \pm 0.095$ & $-0.702 \pm 0.093$ & $-0.828 \pm 0.024$ & $0.735 \pm 0.028$ & $-1.420 \pm 0.015$ \\
Subset 2 & $0.206 \pm 0.071$ & $-0.306 \pm 0.123$ & $-0.819 \pm 0.112$ & $-0.839 \pm 0.034$ & $0.693 \pm 0.040$ & $-1.408 \pm 0.022$ \\
		\bottomrule
	\end{tabulary}
	\label{tab:repeatability}
\end{table}

\subsection{Mapping Accuracy}
\begin{figure}[H]
	\centering
	\begin{subfigure}{0.49\textwidth}
		\includegraphics[width=1.\textwidth]{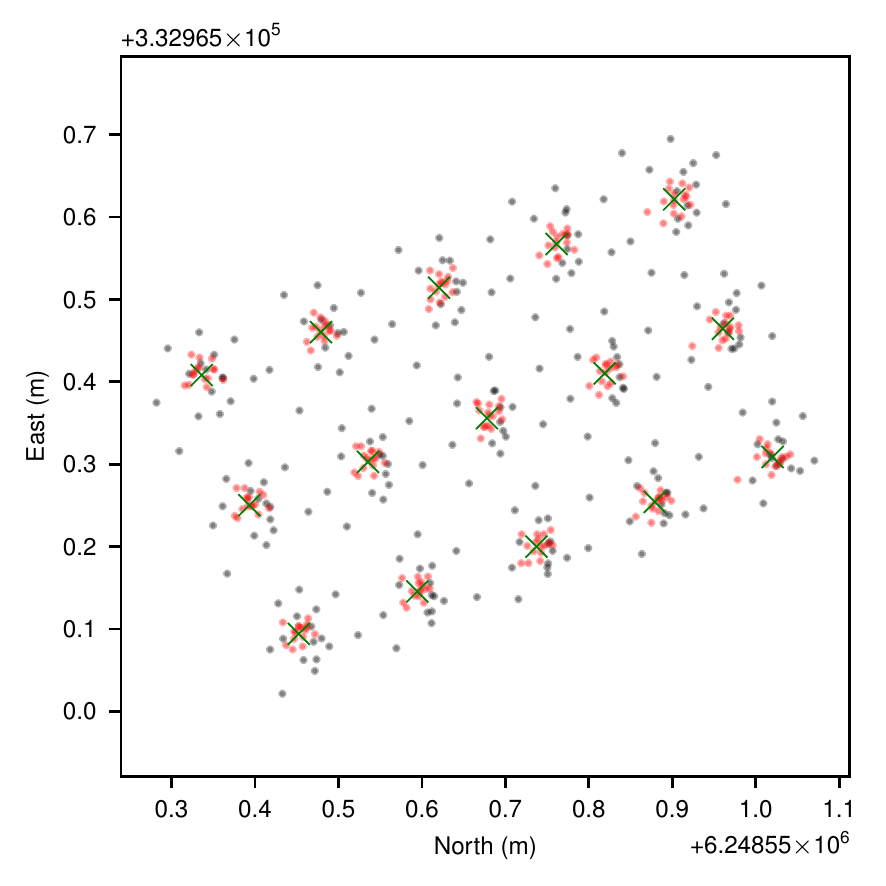}
		\caption{Ladybird - all projections}
		\label{fig:all_projections_ladybird}
	\end{subfigure}
	\begin{subfigure}{0.49\textwidth}
		\includegraphics[width=1.\textwidth]{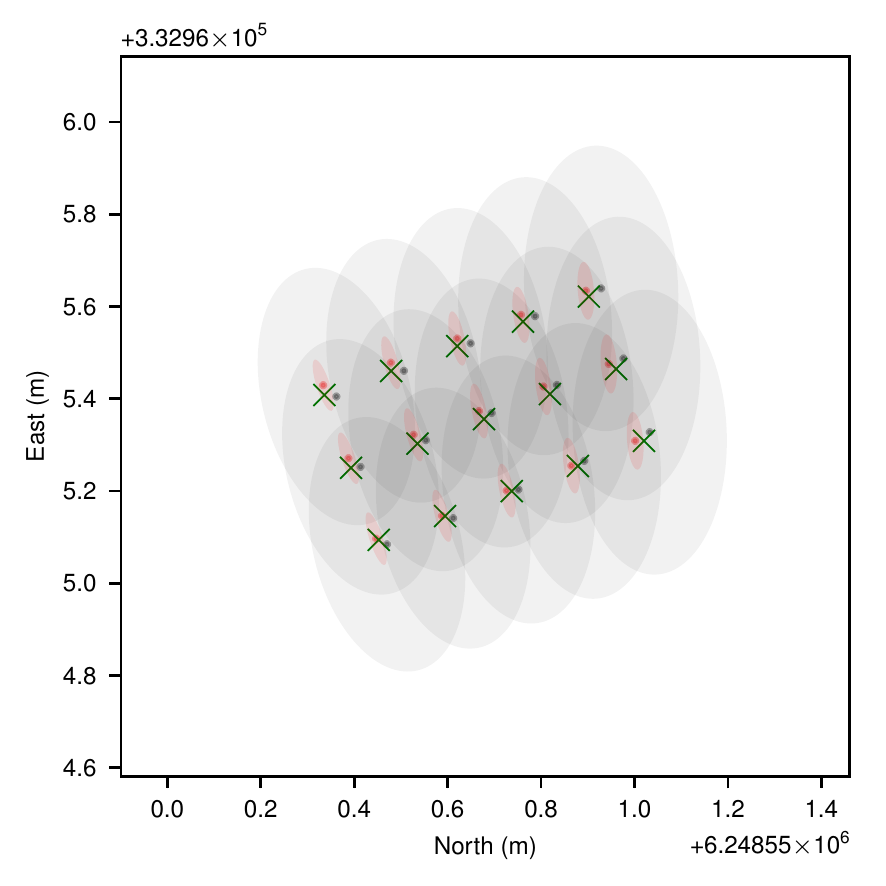}
		\caption{Ladybird - sample projection and uncertainties}
		\label{fig:sample_projection_ladybird}
	\end{subfigure}
	\begin{subfigure}{0.49\textwidth}
		\centering
		\includegraphics[width=1.\textwidth]{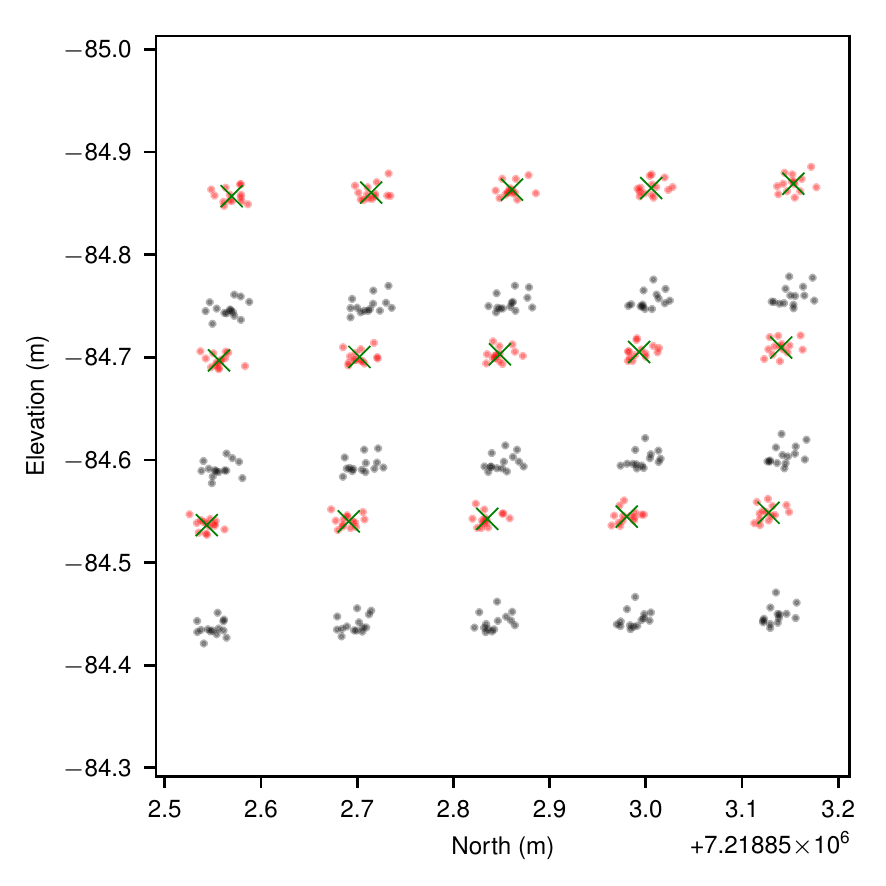}	
		\caption{Shrimp - all projections}
		\label{fig:all_projections_shrimp}
	\end{subfigure}
	\begin{subfigure}{0.49\textwidth}
		\centering
		\includegraphics[width=1.\textwidth]{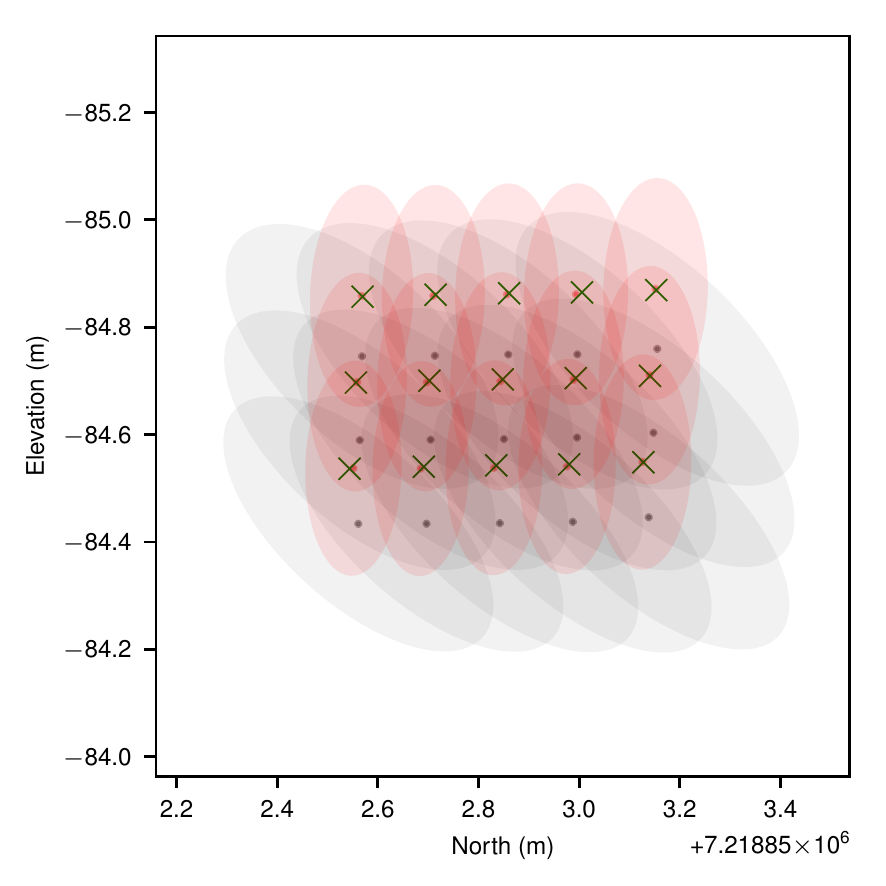}
		\caption{Shrimp - sample projection and uncertainties}
		\label{fig:sample_projection_shrimp}
	\end{subfigure}
	\vspace{-12pt}
	\caption{Point to plane projection comparisons for both platforms. In (\textbf{a},\textbf{c}) all scans of all pattern points are plotted to best fit planes before (grey) and after (red) calibration, while in (\textbf{b},\textbf{d}) a single projected observation of each pattern point is shown with 1-$\sigma$ uncertainty ellipses. In all figures, post-calibration pattern point estimates are marked with green crosses for reference. Because the calibration pattern was positioned flat on the ground for Ladybird, a top down (North-East) view was selected for (\textbf{a},\textbf{b}). Conversely, for Shrimp the calibration pattern was positioned almost vertically on a ladder, facing west, and therefore a side view (North-Elevation) was selected for (\textbf{c},\textbf{d}).} 
	\label{fig:projection_comparison}
\end{figure}

Figure \ref{fig:projection_comparison} shows Ladybird's and Shrimp's observations, after outlier removal, projected to the best fit plane of predicted point locations before and after calibration (see Section \ref{sec:method_projection}). Figure \ref{fig:projection_comparison}a,c give projections of all observations, while Figure \ref{fig:projection_comparison}b,d only show the projections for one observation, but include uncertainty ellipses. Post calibration average pattern point estimates are also shown as green crosses. 

For both platforms, the calibrated camera pose exhibited more densely bound projected points. The change in spread is less pronounced for the Shrimp platform, because as mentioned previously the manual measurements were by chance much closer to the optimum values, though a significant offset can be observed between manual and calibrated results. The plots also demonstrate the effect of relative camera pose uncertainty on mapping uncertainty, which were significant for both Ladybird and Shrimp.

\begin{figure}[H]
	\centering
	\includegraphics[width=0.49\textwidth]{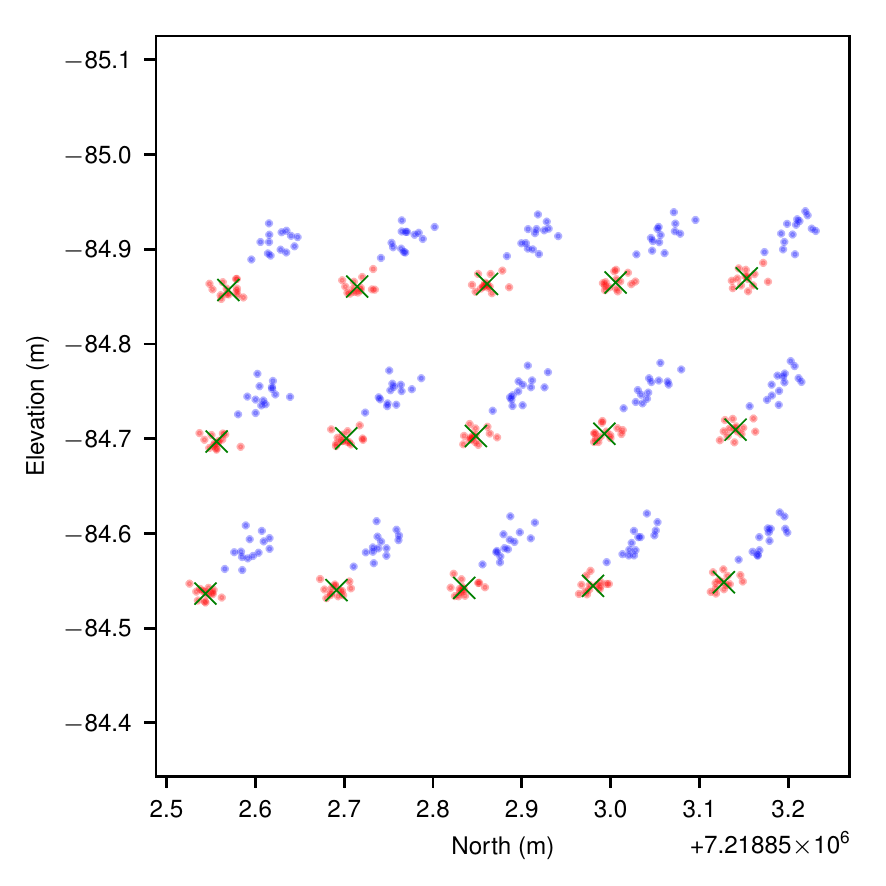}	
	\vspace{-12pt}
	\caption{\textls[-25]{All pattern points plotted to best fit planes for Shrimp, red for post calibration (same as in \ref{fig:all_projections_shrimp})} and blue for projected points resulting from a relative camera pose where the optimised orientation was altered by 1\degree. This demonstrates the significant effect small changes in the camera pose can have on mapping performance.}
	\label{fig:shrimp_projection_1deg_diff}
\end{figure}

Given the hand measurement for shrimp happened to be close to the optimal, the effect of adding just one degree of error to the measured axis-angle vector is shown in Figure \ref{fig:shrimp_projection_1deg_diff}, which compares the mapped points from the optimised and erroneous camera pose. The optimisation improves the cluster significantly compared to a hand measured pose with one degree error.

\subsection{Basin of Attraction} \label{sec:boa}
\begin{figure}[H]
	\centering
	\begin{subfigure}{0.49\textwidth}
		\centering
		\includegraphics[width=1.\textwidth]{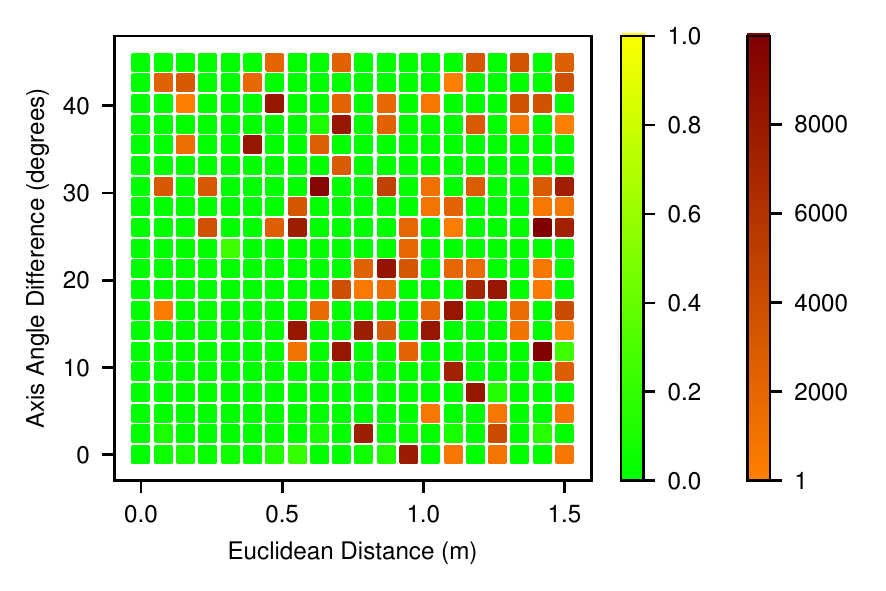}	
		\caption{Ladybird}
		\label{fig:boa_ladybird}
	\end{subfigure}
	\begin{subfigure}{0.49\textwidth}
		\centering
		\includegraphics[width=1.\textwidth]{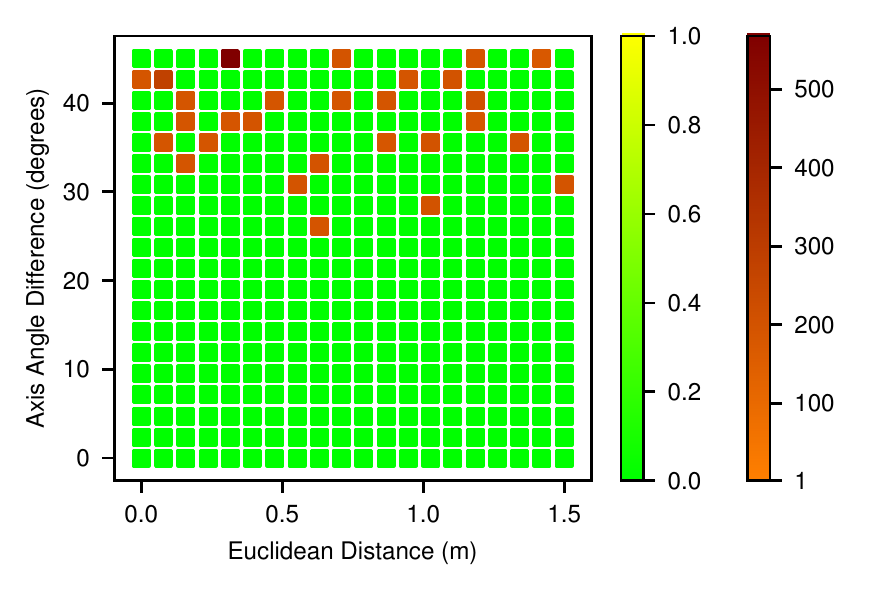} 
		\caption{Shrimp}
		\label{fig:boa_shrimp}
	\end{subfigure}
	\vspace{-12pt}
	\caption{Basins of attractions for the (\textbf{a}) Ladybird and (\textbf{b}) Shrimp platforms. The x and y axes of the plots are the Euclidean distances and axis-angle rotational differences between the initial values and the optimal reference solution respectively. Each grid cell is colour coded into two tiers based on the Mahalanobis distance of the result to the known optimum: below 0.1 and above, which was chosen as a suitable threshold for optimisation success.} 
	\label{fig:basins_of_attraction}
\end{figure}

Basin of attraction plots, which were generated as described in Section \ref{sec:method_basin_of_attraction} using the Powell optimiser, are shown in Figure \ref{fig:basins_of_attraction} for both platforms. The Mahalanobis distances were generally either close to zero or very large, so they are colour coded into two tiers, below and above 1.0, to improve readability. The basin for the Ladybird platform (Figure \ref{fig:boa_ladybird}) shows success can be expected in the triangular region with less than 20\degree \,and 0.5 m of hand measurement error. The basin for Shrimp (Figure \ref{fig:boa_shrimp}) shows a greater immunity to translation errors and successful results can be expected with initialisation errors less than 20\degree \,and as high as 1.5 m. Both figures indicate that when the initialisation error is higher, there is still a high chance (approx. 60\%) of a solution within a Mahalanobis distance of 0.1, but it cannot be relied upon, and deteriorates as the distance from optimum increases.

\section{Discussion} \label{sec:discussion}
The results show that the proposed method was able to to reliably estimate the relative line scan camera pose on a mobile ground vehicle, resulting in a reduction in mapping error, as long as the calibration data includes sufficient viewpoint variability. An uncertainty of 0.06 m/0.018 rad (1.05\degree) for Ladybird, and 0.18 m/0.042 rad (2.39\degree) for Shrimp was achieved. This result is dependent on the certainty of input parameters, which include pixel observations, navigation system solutions, and camera intrinsics. For example, confidence in the calibrated pose parameters for the Ladybird platform was significantly greater than for the Shrimp platform, due to Ladybird's higher grade \ac{IMU}, which allowed the navigation system to provide more certain solutions.

The results show that it is necessary to examine reprojection errors and remove outliers, as is common with many camera calibration approaches. Outliers statistically fall outside the assumptions encoded in their respective error models, and so the mean of the final camera pose distribution is pulled in the wrong direction. A number of observations for both platforms exhibited high reprojection errors relative to other observations (>approx. 8 pixels). These errors could be caused by manual labelling inaccuracies (e.g., due to limited resolution), navigation system solution errors that incorrectly fall outside the reported navigation uncertainty, or a combination thereof. Removing outliers had significant effects on the results, evident particularly in the correction of some parameters such as the z-offset $\mathbf{t}_{c,z}^b$ for both platforms (see Table \ref{tab:results}). The results shown in Figure \ref{fig:outlier_removal} also support the iterative removal of outliers at each stage. For instance, reprojection errors of Ladybird observation 20~improve as other observations are removed, while Shrimp observation 4 degrades, and was eventually removed. Conversely, as shown in Figure \ref{fig:uncertainty_increase} a larger number of observations, outliers or not, allows for greater certainty in the final result. Thus, there are two competing factors when performing outlier rejection. A sufficiently large number of observations is required to maintain an acceptable level of certainty, yet removing outliers is important to minimise reprojection errors. It is, therefore, desirable to obtain a sufficient number of observations to allow for subsequent outlier removal. The additional computational time required when increasing the number of observations is also a consideration. The~ending condition threshold should be chosen such that significant outliers are removed, but a sufficient number of observations remain. In this paper a value of 5 pixels was empirically determined as an appropriately balanced threshold given the data.

The main product of the \ac{MCMC} uncertainty analysis is a covariance matrix (Equation (\ref{eq:camera_pose_covariance})), which can be used to estimate mapping accuracy (e.g., Figure \ref{fig:projection_comparison}a,c). However, covariance matrices represent uncertainty in a compressed format, given the assumption that the likelihood function is normally distributed. The corner plots (Figure \ref{fig:corner_plot})  provide a direct view of the \ac{MCMC} sampling result, which qualitatively confirm that for both vehicles the normality assumption is justified: specifically that the distributions behave linearly within the sampled region, and the 1D histograms are qualitatively Gaussian in shape.  

In this paper we propose methods of visualising sensor pose distributions in a human interpretable way, as depicted in Figures \ref{fig:spheres} and \ref{fig:result_on_model}. The sphere plots and associated axis-angle magnitude histograms in Figure \ref{fig:spheres} present the same underlying data in the MCMC sample plots (Figure \ref{fig:corner_plot}), but focus on human interpretability of the orientation parameters. The sphere provides a relatable reference that demonstrates how closely clustered the pose orientation is. Similarly, the visualisation in Figure \ref{fig:result_on_model} allows for human interpretation of the resulting camera pose and uncertainty (likelihood distribution) with respect to the platform models. These figures particularly highlight the greater uncertainty in translation parameters compared to orientation. They also confirm that the solutions are qualitatively ``sensible'' with respect to the physical platform models. 

The primary objective of optimising the camera pose is to reduce mapping errors. This was demonstrated in the results by the tighter clustering of mapped calibration target points that was achieved post-calibration. The improvement was particularly noticeable for the Ladybird platform, and to a lesser extent for the Shrimp platform. By chance, the manually measured camera pose on Shrimp was much closer to the optimal result than it was for Ladybird, and so the mapping improvement for Shrimp was less pronounced. The camera location on the Shrimp platform was easier to access, due to the lower height and smaller footprint, compared to the Ladybird platform, which likely explains the better manual estimate. Nevertheless, such accurate manual measurements can typically not be guaranteed, and Figure \ref{fig:shrimp_projection_1deg_diff} reveals the sensitivity of the map to small errors in camera pose, highlighting the need for calibration.

The results reinforce the importance of acquiring a calibration dataset that exhibits a wide variety of platform poses with respect to the calibration pattern. This was tested by optimising both with and without large body roll ($\phi_{b,x}^w$) observations. Removing high $\phi_{b,x}^w$ observations had a considerable effect, as shown in Table \ref{tab:angular_diversity}, where uncertainty approximately doubled and even tripled for some parameters. 

The proposed method is able to deal with a wide range of initial hand measured values when paired with the Powell optimisation algorithm. In Figure \ref{fig:basins_of_attraction} we propose an intuitive approach to visualising the basins of attraction, by reducing the 6 \ac{DOF} initial parameter space to form a 2D pose-distance space, comprising Euclidean and axis-angle distance. The plots demonstrate that initial estimates that deviate up to 0.5 m or 20\degree \,are likely to result in a successful optimisation. Additionally, with even larger deviations there is still a better than even (~60\%) chance of success. However, this is highly dependent on the geometry of the sensor and platform, the acquired data and the chosen optimisation algorithm. In our case, the solution for the Shrimp platform was surprisingly robust to initial translation errors (up to 1.5 m), while some failures can be seen at over 0.5 m for the Ladybird platform. This may be the result of the greater platform roll and pitch angles (up to 17\degree) achieved with Shrimp during data acquisition. In addition, different optimisers will have varying abilities to deal with local minima and ``flat'' regions in the 6 \ac{DOF} parameter space.  Nevertheless, measurement error tolerances of $\pm$0.5 m and $\pm$20\degree \,should be practically achievable for most applications.

As shown in Table \ref{tab:repeatability}, the camera pose estimates obtained from two independent datasets for the \textls[-5]{same platform were consistent with each other. The results therefore verify that the proposed approach is repeatable by demonstrating that different sets of data from the same platform yield consistent results.}   

An important advantage of the proposed method is that exact dimensions of the calibration pattern do not need to be known. As such, a planarity assumption or assumptions about the distances or geometry between points are not required. This simplifies the method in the field because it is not affected by printing errors or damage/warping of the pattern which affects the relative geometry of the points. Furthermore, a calibration pattern could be manually produced in the field if necessary. One important condition, however, is that individual points are uniquely distinguishable in the line scan image data.

\section{Conclusions}
This paper demonstrated a novel method for estimating a rigidly mounted line scanning camera's fixed 6 \ac{DOF} pose relative to a mobile platform with a navigation system. The method is appropriate for ground or very low altitude applications, where the scene is relatively near the platform, as it does not require \acp{GCP} and uses a compact calibration pattern, the dimensions of which do not need to be known. Furthermore, it does not require data from auxiliary sensors such as full frame cameras. The approach involves imaging a calibration pattern with distinctly identifiable points from various platform poses, and using the navigation system and image data to triangulate their positions in the world frame. Reprojecting the points to the camera yields reprojection errors, which are used as a basis for outlier rejection, and then to calculate the likelihood given a candidate camera pose. By~minimising the negative log likelihood, the optimal relative camera pose can be obtained. Given~the likelihood function, an \ac{MCMC} algorithm is able to estimate the certainty of the camera pose. The~results demonstrate the effectiveness of the approach using two different mobile platforms with differing mounting configurations. The method was shown to be robust to relatively inaccurate initial hand measurements (within 0.5 m and 20\degree). Additionally, a number visualisations have been proposed to aid in human interpretation of the results. Future work will attempt to precisely specify platform pose requirements prior to data collection, automate and improve the pattern point extraction process, and explore the application of a robust optimisation routine or loss function to simplify the outlier rejection process.

\vspace{6pt} 

\acknowledgments{This work is supported by the Australian Centre for Field Robotics (ACFR) at The University of Sydney and by funding from the Australian Government Department of Agriculture and Water Resources as part of its Rural R\&D for profit programme. For more information about robots and systems for agriculture at the ACFR, please visit \url{http://sydney.edu.au/acfr/agriculture}.}

\authorcontributions{Both A.W. and J.U. gathered the relevant data in the field. A.W. conceived and implemented the method, while J.U. supervised the work and provided significant conceptual input. The~paper was written by A.W. and reviewed by J.U.}

\conflictsofinterest{The authors declare no conflict of interest. The founding sponsors had no role in the design of the study; in the collection, analyses, or interpretation of data; in the writing of the manuscript, and in the decision to publish the results.} 

\reftitle{References}
\externalbibliography{yes}

\end{document}